\documentclass[11pt]{article}

\usepackage{arxiv}

\usepackage[utf8]{inputenc}
\usepackage[T1]{fontenc}
\usepackage{lmodern}
\usepackage{microtype}
\usepackage{amsmath,amsfonts}
\usepackage{graphicx}
\usepackage{booktabs}
\usepackage{longtable}
\usepackage{enumitem}
\usepackage{float}
\usepackage{url}
\usepackage[numbers,sort&compress]{natbib}

\usepackage[colorlinks=true,allcolors=blue]{hyperref}

\title{Seeing Soil from Space:\\Towards Robust and Scalable\\Remote Soil Nutrient Analysis}

\author{
 David Șeu\thanks{Corresponding author: david.seu@co2angels.com. 
 Work partially conducted while the author was a visiting researcher at the ESA \(\Phi\)-Lab.} \\
  CO2 Angels\\
  European Space Agency \(\Phi\)-Lab\\
  Cluj-Napoca, Romania \\
  \texttt{david.seu@co2angels.com} \\
   \And
  Nicolas Longépé\\
  European Space Agency \(\Phi\)-Lab\\
  Frascati, Italy \\
  \texttt{Nicolas.Longepe@esa.int} \\
  \And
  Gabriel Cioltea \\
  CO2 Angels\\
  Cluj-Napoca, Romania \\
  \texttt{gabriel.cioltea@co2angels.com} \\
    \And
  Erik Maidik \\
  CO2 Angels\\
  Cluj-Napoca, Romania \\
  \texttt{erik.maidik@co2angels.com} \\
    \And
  Călin Andrei \\
  CO2 Angels\\
  Cluj-Napoca, Romania \\
  \texttt{calin.andrei@co2angels.com} \\
}

\begin{document}
\maketitle
\begin{abstract}
Environmental variables are increasingly affecting agricultural decision-making, yet accessible and scalable tools for soil assessment remain limited. This study presents a robust and scalable modeling system for estimating soil properties in croplands, including soil organic carbon (SOC), total nitrogen (N), available phosphorus (P), exchangeable potassium (K), and pH, using remote sensing data and environmental covariates. The system employs a hybrid modeling approach, combining the indirect methods of modeling soil through proxies and drivers with direct spectral modeling. We extend current approaches by using interpretable physics-informed covariates derived from radiative transfer models (RTMs) and complex, nonlinear embeddings from a foundation model. We validate the system on a harmonized dataset that covers Europe's cropland soils across diverse pedoclimatic zones. Evaluation is conducted under a robust validation framework that enforces strict spatial blocking, stratified splits, and statistically distinct train-test sets, which deliberately make the evaluation harder and produce more realistic error estimates for unseen regions. The models achieved their highest accuracy for SOC and N. This performance held across unseen locations, under both spatial cross-validation and an independent test set. SOC obtained a MAE of 5.12 g/kg and a CCC of 0.77, and N obtained a MAE of 0.44 g/kg and a CCC of 0.77. We also assess uncertainty through conformal calibration, achieving 90\% coverage at the target confidence level. This study contributes to the digital advancement of agriculture through the application of scalable, data-driven soil analysis frameworks that can be extended to related domains requiring quantitative soil evaluation, such as carbon markets.
\end{abstract}

% keywords can be removed
%\keywords{First keyword \and Second keyword \and More}

\section{Introduction}

Agriculture is the foundation of global food security, with soil as a key resource for production \cite{Kopittke2019EnvInt}. In the 21st century, agriculture continues to rely mainly on traditional practices with comparatively low levels of digitalization \cite{Choruma2024JAFR}. This limited digitalization is particularly concerning given the increasing challenges the sector faces. Global food demand is projected to increase by \mbox{35--56\%} by 2050 relative to 2010, making future production more difficult to meet under land and climate constraints \cite{vanDijk2021NatFood, FAO2023SOFI}. Across major crops, yield growth has stagnated under climate variability \cite{Ray2013PLOSONE, Lobell2011Science, GAPReport2023}. Agriculture, forestry, and other land use account for \mbox{$\sim$13--21\%} of anthropogenic greenhouse gas emissions, and soil degradation is widespread, with an estimated one-third of global soils moderately to severely degraded \cite{Smith2024, IPCC2022WGIII, FAO2015SWSR}. In particular, nutrient depletion and declines in soil organic matter reduce soil fertility, water-holding capacity, and aggregate stability, thereby affecting long-term productivity and resilience \cite{Srivastava2024}. Addressing these challenges requires agricultural systems to simultaneously increase productivity, maintain farmer profitability, and reduce negative environmental impacts. Achieving this "triple challenge" depends on more efficient input management.

Fertilizers, in particular, represent one of the highest operational costs for farms and are essential to sustaining yields \cite{USDAERS2023FertilizerCosts}. Recent isotopic tracing studies report 21--44\% of applied nitrogen fertilizer is taken up by crops, with the remainder being lost to leaching, volatilization, or denitrification \cite{,Xu2025, Govindasamy2023}. These inefficiencies impose economic losses, degrade soil health, and contribute to greenhouse gas emissions. Soil sampling remains the conventional method for fertilizer planning, but adoption is constrained by cost, time, and logistical barriers \cite{OConnell2022Geodrs}. Although precision agriculture technologies exist, their uptake remains limited due to high costs, technical complexity, and gaps in spatial data coverage \cite{Gebbers2010, Finger2019}.

Recent policy developments further highlight the urgency of scalable soil monitoring. The European Union’s Soil Monitoring Law establishes a harmonized framework to evaluate soil health across member states, with indicators such as the SOC-to-clay ratio \cite{eu_soil_monitoring_2023}. As more than 60\% of EU soils are classified as unhealthy, scalable approaches are required to generate timely and field-specific soil data to support both farm productivity and environmental sustainability \cite{Chowdhury2024DiB, Panagos2025GlobalChallenges}.

However, monitoring soil health at scale is not trivial. Remote sensing provides proxies for soil conditions, but direct detection of most soil chemical properties remains constrained by vegetation cover, shallow penetration, and low signal-to-noise ratios in relevant spectral domains \cite{angelopoulou2019remote, koumoulidis2024review, Salam}. Even in bare-soil conditions, reflectance is typically limited to the uppermost soil layer \cite{Norouzi2021}. Unlike broad trend analyses for agronomic decision-making, robust systems require harmonized ground truth, transparent validation, and explicit uncertainty estimates to ensure practical reliability.

In this study, we present a scalable and robust system to estimate soil properties relevant for agronomic management, including SOC, total nitrogen (N), available phosphorus (P), exchangeable potassium (K), and soil pH (in H\textsubscript{2}O), on croplands. The system integrates Earth Observation (EO) data and environmental covariates using machine learning techniques to provide three-dimensional plus time (3D+T) estimates, capturing spatial variability, depth profiles, and temporal dynamics. This work contributes through its scale and design, notably the integration of physics-informed machine learning with state-of-the-art deep learning approaches, and the development of a robust and transparent validation framework. The objective is to evaluate the performance and limitations of such a system for agronomic decision-making, with a focus on enabling more precise and sustainable fertilizer management.

\section{Materials \& Methods}
\label{sec:material_methods}

The proposed methodological framework builds on the SCORPAN paradigm of soil science \cite{McBratney2003}, which conceptualizes soil properties as a function of state factors,
\[
S = f(c,o,r,p,a,n,t),
\]
namely climate (\(c\)), organisms (\(o\)), relief (\(r\)), parent material (\(p\)), age (\(a\)), spatial position (\(n\)), and time (\(t\)). This formalism provides the basis for digital soil mapping: we instantiate \(f(\cdot)\) with covariates, separating static factors (e.g., long-term climate normals, terrain, soils, geomorphology, land cover) from dynamic factors (e.g., monthly climate forcing, gap-free multispectral reflectance, phenology and productivity proxies, physics-informed canopy traits, and bare-soil reflectance). This separation makes the temporal factor explicit, modeling soil properties as a function of both static state factors and dynamic temporal covariates across multiple scales.
The modeling objective is then to learn spatio–temporal mappings from these covariates to target soil properties while enforcing robustness and interpretability across heterogeneous landscapes.

In this study, we extend the conventional SCORPAN framework with a hybrid modeling strategy that combines physics-informed spatio-temporal machine learning \cite{Karniadakis2021, Reichstein2019, meng2025physics} with foundation models \cite{Reichstein2023}. Physics-informed components embed environmental constraints, improving interpretability, while foundation models leverage large-scale representation learning to enhance predictive performance. The modeling operates at the pixel level, which allows efficient scaling to continental extents while simultaneously preserving local variability. 

To ensure reliability, we employ an evaluation framework that explicitly accounts for dataset heterogeneity, with particular emphasis on AEZ variation. In addition to standard accuracy metrics, we incorporate uncertainty quantification via conformal prediction \cite{tseng2024lightweightpretrainedtransformersremote} to report probabilistic coverage. This enables assessment not only of predictive performance but also of confidence, especially in underrepresented environments.

The following subsections describe the soil datasets used as ground truth (Section~\ref{sec:study_area_soil_sample_data}), the static and dynamic covariates employed to represent environmental state factors (Section~\ref{sec:env_covariates}), and the modeling and evaluation framework, including strategies for uncertainty quantification (Section~\ref{sec:modeling_approach}).

\subsection{Study Area \& Soil Sample Data}
\label{sec:study_area_soil_sample_data}

\begin{figure}[!htbp]
    \centering
    \includegraphics[width=0.9\textwidth]{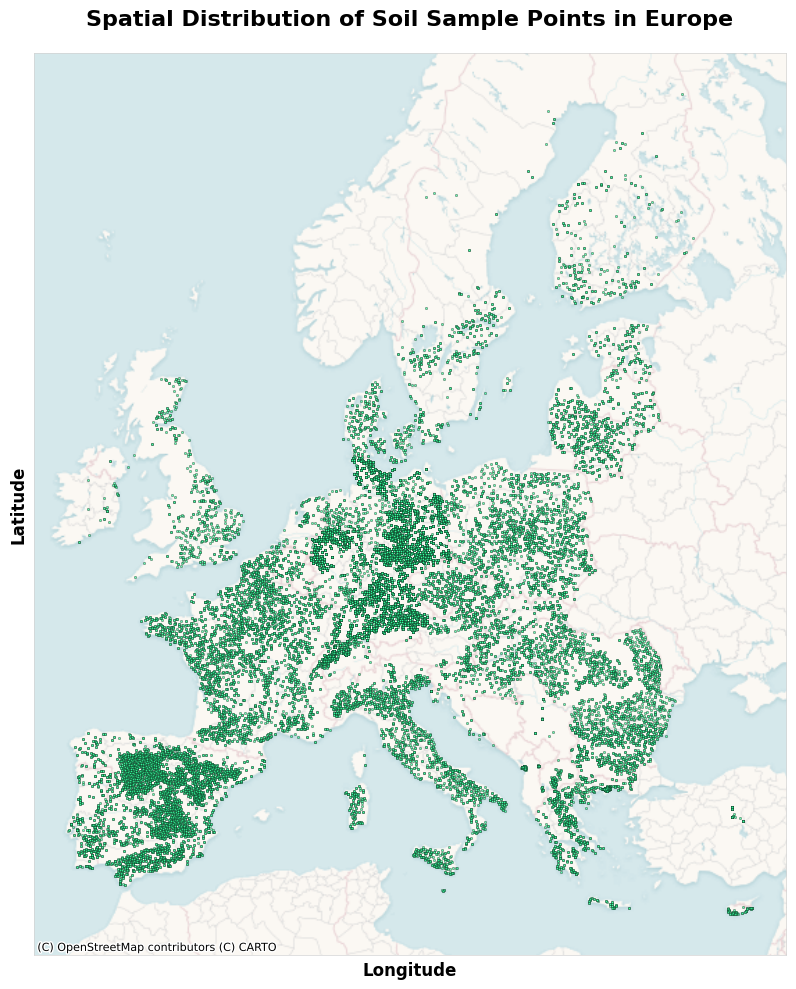}
    \caption{Spatial distribution of soil samples included in this study.}
    \label{fig:space_points}
\end{figure}

This study is based primarily on the Land Use and Land Cover Survey (LUCAS) topsoil surveys, the largest harmonized soil dataset for Europe \cite{Orgiazzi2018EJSS, Jones2020LUCAS2015, JRC2020LUCAS2018}. To broaden coverage, we harmonized additional regional and national surveys to the LUCAS analytical framework, resulting in a comprehensive representation of cropland soils across diverse pedoclimatic zones. We retained only cropland points, and outliers were removed per variable using domain-expert review. The final harmonized dataset comprises 31{,}904 samples, which contain the point location (latitude and longitude), year of sample, and target values (Figure~\ref{fig:space_points}).

\begin{figure}[!htbp]
    \centering
    \includegraphics[width=1\textwidth]{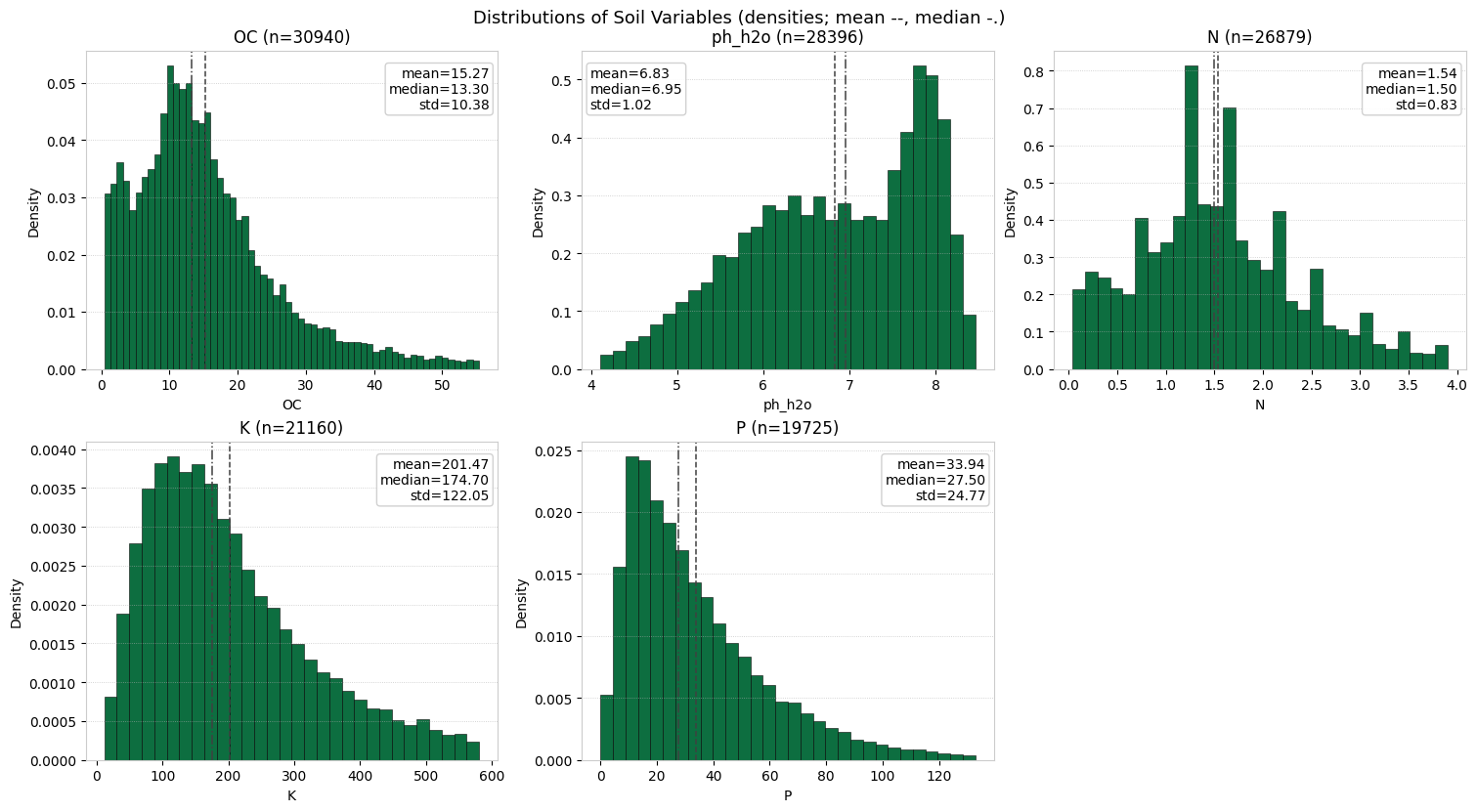}
    \caption{Histograms of SOC, N, P, K, and pH across the harmonized dataset.}
    \label{fig:soil_histograms}
\end{figure}

The soil variables show pronounced heterogeneity (Figure~\ref{fig:soil_histograms}). SOC, P, and K are right-skewed, while N and pH are multimodal, reflecting variation across agro-ecological contexts. When disaggregated by AEZs, clear zone-specific nutrient patterns emerge (Table~\ref{tab:aez_major_stats}): SOC and N are elevated in cold and terrain-limited zones but lower in irrigated or subtropical soils, pH ranges from acidic in temperate–cool zones (median $\sim$6.6) to alkaline in subtropics ($\sim$7.7--7.8), P peaks in temperate–cool zones ($\sim$35 mg/kg) but declines in subtropics ($\sim$16--19 mg/kg), while K is highest in subtropics ($\sim$200 mg/kg) and lowest in cold zones ($\sim$82 mg/kg). These gradients underscore how nutrient values are closely tied to AEZs.

\clearpage

\begin{figure}[!htbp]
    \centering
    \includegraphics[width=0.8\textwidth]{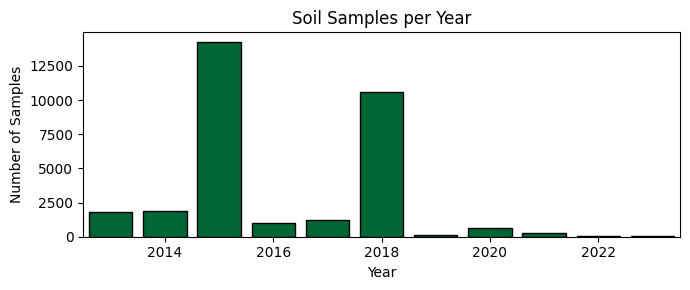}
    \caption{Temporal distribution of soil samples by year of collection.}
    \label{fig:soil_years}
\end{figure}

The dataset is also imbalanced. Temperate–cool zones dominate with over 17{,}000 samples, whereas cold, terrain-limited, or subtropical areas contribute only a few hundred to a few thousand samples. Depth categories are similarly skewed, with the majority of samples collected in the 0--30 cm range (over 26{,}000), while deeper horizons (30--60 cm and 60+ cm) are far less represented (Table~\ref{tab:depth_dist}). Such imbalances risk underrepresenting minority environments and soil layers, and necessitate stratified evaluation (Table~\ref{tab:aez_major_stats}). By contrast, temporal coverage is uneven but less critical: most samples come from campaigns in 2015 and 2018 (more than 70\%), while samples from other years are sparsely represented (Figure~\ref{fig:soil_years}). Because soil nutrients change slowly over multiannual timescales \cite{Seabloom2021SoilCarbon}, spatial heterogeneity, AEZ imbalance, and depth distribution remain the dominant sources of bias, while temporal concentration mainly reflects survey logistics.

\begin{table}[!htbp]
\centering
\caption{Depth category distribution.}
\label{tab:depth_dist}
\begin{tabular}{lccc}
\toprule
 & 0--30 cm & 30--60 cm & 60+ cm \\
\midrule
Count & 26,494 & 2,197 & 3,213 \\
\bottomrule
\end{tabular}
\end{table}

\begin{table}[!htbp]
\centering
\small
\caption{Median [Interquartile Range] of soil variables by major AEZ family.}
\label{tab:aez_major_stats}
\begin{tabular}{lrrrrrr}
\toprule
AEZ family & $n$ & SOC (g/kg) & N (g/kg) & pH$_{\mathrm{H_2O}}$ & P (mg/kg) & K (mg/kg) \\
\midrule
Temperate — cool          & 17{,}390 & 13.80 [13.00] & 1.50 [1.20] & 6.61 [1.50] & 34.80 [32.80] & 156.90 [142.70] \\
Irrigated/Hydromorphic    & 6{,}271  & 12.34 [11.17] & 1.50 [0.90] & 7.35 [1.51] & 27.60 [33.10] & 182.40 [174.15] \\
Subtropics — cool        & 3{,}791  & 12.00 [11.10] & 1.30 [0.90] & 7.83 [1.06] & 18.70 [18.68] & 198.85 [174.65] \\
Subtropics — mod.\ cool  & 2{,}551  & 12.20 [9.10]  & 1.30 [0.80] & 7.74 [1.38] & 15.70 [18.00] & 209.10 [216.70] \\
Temperate — moderate      & 1{,}030  & 15.55 [8.60]  & 1.70 [0.70] & 6.89 [1.67] & 15.90 [17.80] & 208.80 [130.00] \\
Terrain/land limits       &   695    & 19.33 [17.58] & 1.95 [1.30] & 7.36 [1.70] & 25.60 [29.98] & 191.50 [181.20] \\
\bottomrule
\end{tabular}
\end{table}

Crop management and crop type also shape nutrient distributions through differences in fertilization, residue management, and rotations \cite{Mukhametov2024CropRotation, Jiang2022SoilHealth}. For example, perennial systems tend to accumulate more SOC than annual cereal-based systems \cite{Bergquist2025SoilCarbon, Siddique2023Perennialization}. However, crop type information is not consistently available across the integrated datasets, limiting our ability to analyze this dimension. LUCAS surveys are among the only ones that provide crop type labels, preventing a systematic crop-specific assessment across Europe.

Together, these results highlight that nutrient distributions are shaped by agro-ecological conditions, and that both skewed distributions and spatial imbalance must be explicitly addressed in model design and evaluation.

\subsection{Environmental Covariates}
\label{sec:env_covariates}

\subsubsection{Static Covariates}

Static, time-invariant covariates were selected to represent the long-term state factors of the SCORPAN paradigm, namely climate (\(c\)), organisms (\(o\)), relief (\(r\)), parent material (\(p\)), and land use (\(n\)). These descriptors provide the environmental baseline against which dynamic drivers act and condition soil nutrient concentrations at decadal to centennial scales.

\textbf{Climate (\(c\)).} Climatologies at High Resolution for the Earth’s Land Surface Areas (CHELSA) bioclimatic variables at 1 km resolution provide 30-year means and extremes of temperature and precipitation. These long-term regimes constrain organic matter turnover and water availability, capturing climatic boundaries that short-term reanalysis cannot resolve  \cite{Karger2017CHELSA}.

\textbf{Agro-ecology (\(o\)).} AEZ descriptors from GAEZ at 5 arcmin resolution delineate suitability classes and growing conditions. By encoding crop–environment adaptation and management intensity, AEZs contextualize soil nutrients within agricultural potential, complementing continuous climatic variables with categorical strata \cite{Fischer2012GAEZ}.

\textbf{Soils (\(p\)).} Global soil type layers from Soil Taxonomy at 250 m resolution capture inherent pedological differences such as clay content, mineralogy, and organic-rich horizons. These intrinsic properties serve as structural baselines, in contrast to vegetation-based proxies that reflect transient states \cite{SoilTaxonomy2014}.

\textbf{Terrain (\(r\)).} Digital elevation models at 30 m resolution provide elevation, slope, aspect, and hydrological derivatives (e.g., topographic wetness index, curvature). These metrics capture redistribution of water and sediment, controlling nutrient leaching, accumulation zones, and effective rooting depth \cite{ho_2025_14900181}. 

\textbf{Geomorphology and lithology (\(p\)).} Landform and parent material classes from the EcoTapestry framework at 250 m resolution encode geological substrates and surface processes. Parent material mineralogy sets long-term P and K supply capacity, linking chemical fertility to geological sources \cite{Hengl2018EcoTapestry}.

\textbf{Land cover (\(n\)).} Cropland extent from Global Land Analysis and Discovery (GLAD) at 30 m resolution and fractional vegetation cover from Global Vegetation Fractional Cover Product (GVFCP) at 1 km resolution provide a representation of land use and canopy cover. These proxies capture management intensity, disturbance history, and the balance between organic matter inputs and decomposition \cite{Thenkabail2021USGS, GVFCP2020}.

\subsubsection{Dynamic Covariates}

Dynamic covariates instantiate the temporal dimension of the SCORPAN paradigm (\(t\)) and its interaction with climate (\(c\)), organisms (\(o\)), and parent material (\(p\)). They are represented as monthly time series for the 24 months preceding soil sampling and capture variability in climate forcing, vegetation function, canopy physiology, and direct soil signals, thus linking aboveground processes to belowground nutrient concentrations. By incorporating physics-informed indices, Radiative Transfer Model (RTM) traits, foundation model embeddings, and bare-soil reflectance, this approach extends standard digital soil mapping to offer explicit temporal structure, physically constrained traits, and high-dimensional embeddings.

\textbf{Climate and energy fluxes (\(c,t\)).} ERA5 reanalysis at $0.25^{\circ}$ resolution \cite{Hersbach2020ERA5} supplied covariates such as monthly temperature and precipitation, complemented by the Clouds and the Earth’s Radiant Energy System (CERES) shortwave fluxes at $1^{\circ}$ resolution \cite{Loeb2018CERES} and Moderate Resolution Imaging Spectroradiometer (MODIS) land surface temperature at 1 km \cite{Wan2015MODISLST}. Together, these variables constrain soil moisture regimes and energy inputs, which govern organic matter turnover and nutrient mobilization.

\begin{figure}[!htbp]
    \centering
    \includegraphics[width=0.9\textwidth]{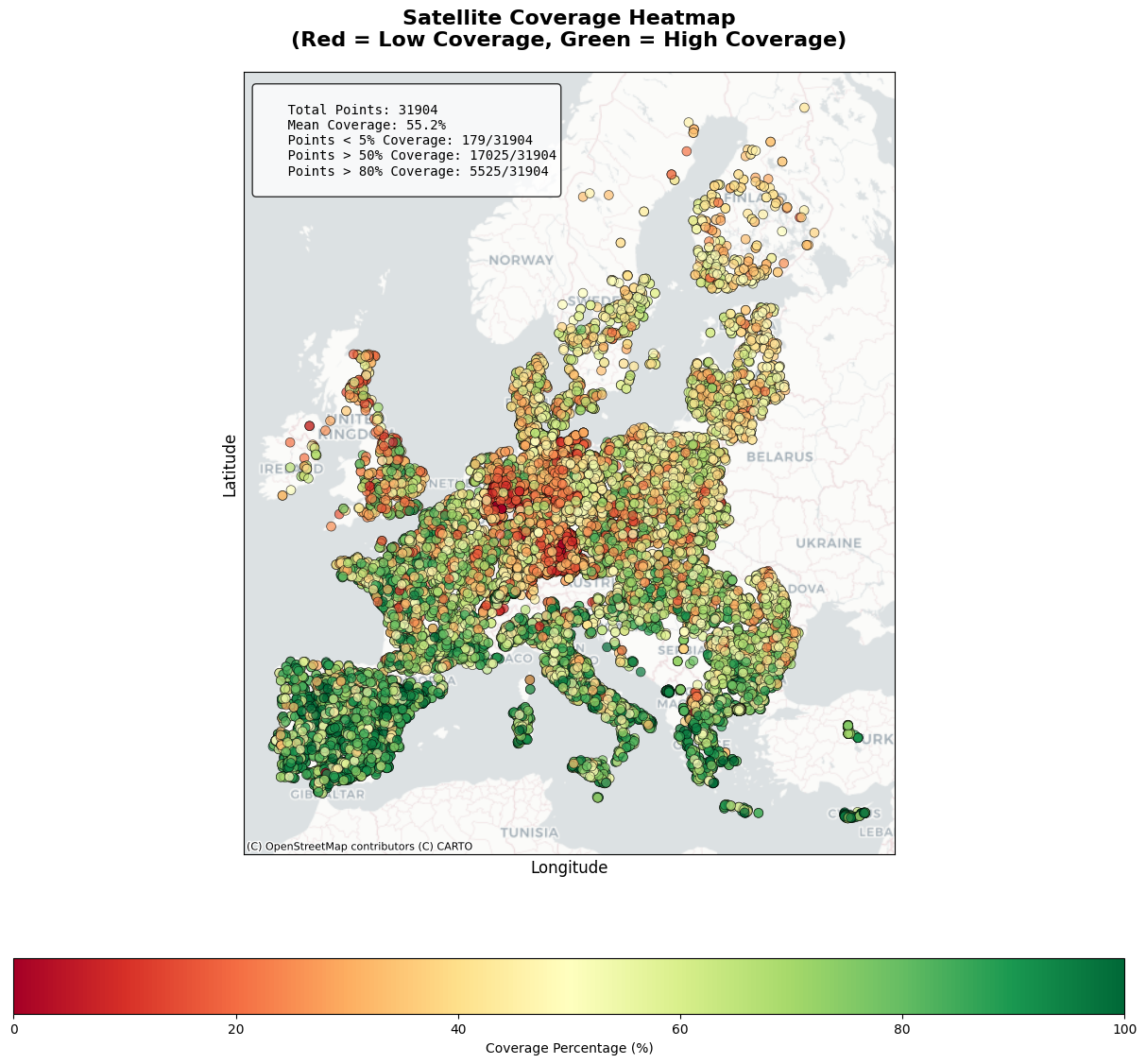}
    \caption{Spatial distribution of valid HLS observations across soil sampling sites after cloud masking. Colors represent the percentage of available monthly reflectance values in the two years prior to sampling. Coverage is heterogeneous across Europe, necessitating fusion and gap-filling to avoid geographic biases in temporal descriptors.}
    \label{fig:coverage}
\end{figure}

\textbf{Gap-free multispectral EO reflectance (\(o,t\)).} Harmonized Landsat--Sentinel (HLS) reflectance (30 m, $\sim$2--3 days) \cite{Claverie2018HLS} and MODIS reflectance (500 m, $\sim$8 days) \cite{Schaaf2021MCD43A4} were cloud-masked and aggregated to monthly composites. To mitigate uneven coverage of HLS across Europe (e.g., persistent gaps in Central Europe and Scandinavia; Figure~\ref{fig:coverage}), we applied fusion and gap-filling strategies, ensuring consistent temporal descriptors \cite{Senty2024}. Without these steps, spatially biased observation density would affect model performance.

\begin{figure}[!htbp]
    \centering
    \includegraphics[width=1\textwidth]{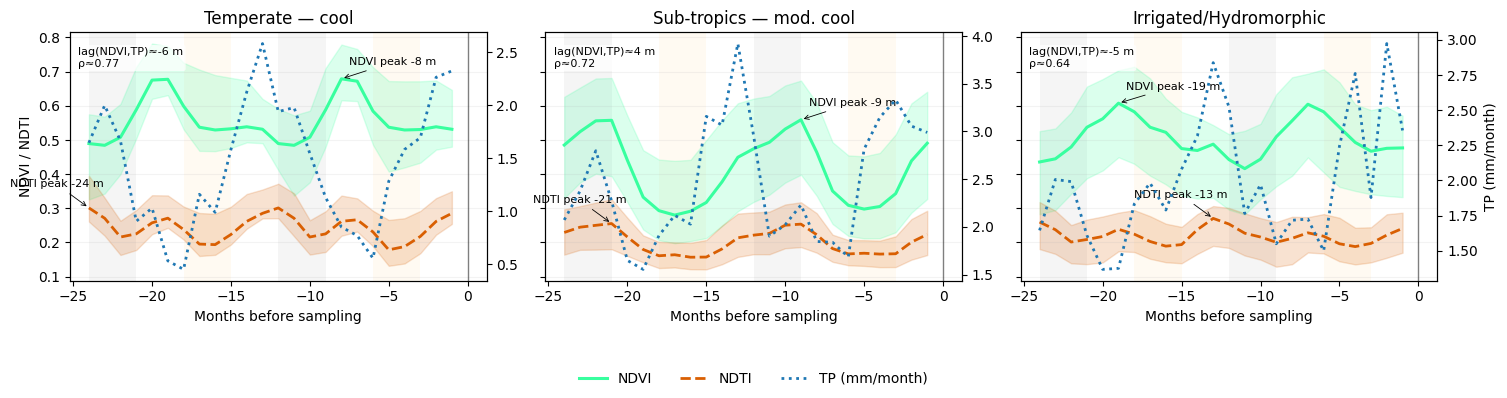}
    \caption{Median and interquartile range of Normalized Difference Vegetation Index (NDVI), Normalized Difference Tillage Index (NDTI), and total precipitation (TP) 24 months before sampling, stratified by AEZ. Temperate systems show regular annual cycles, while subtropical and irrigated systems exhibit water-driven lags and extended greenness. These dynamics represent indirect proxies of nutrient cycling processes.}
    \label{fig:ndvi_sequences}
\end{figure}

\textbf{Vegetation and productivity metrics (\(o,t\)).} From gap-free reflectance, we computed vegetation indices and phenology metrics \cite{Monteith1972, Running2004MOD17}. Seasonal dynamics varied by AEZ (Figure~\ref{fig:ndvi_sequences}): temperate systems showed regular greening and senescence, subtropics displayed rainfall-driven lags, while irrigated and hydromorphic zones sustained extended greenness or decoupled cycles. These descriptors capture processes that directly shape soil nutrient status: peak greenness reflects carbon and N inputs from biomass, senescence marks residue deposition and litter turnover, and soil exposure phases correspond to erosion risk and organic matter mineralization rates. By linking vegetation productivity and soil surface conditions to nutrient cycling, these metrics provide dynamic predictors of soil nutrient status.

\begin{figure}[!htbp]
    \centering
    \includegraphics[width=1\textwidth]{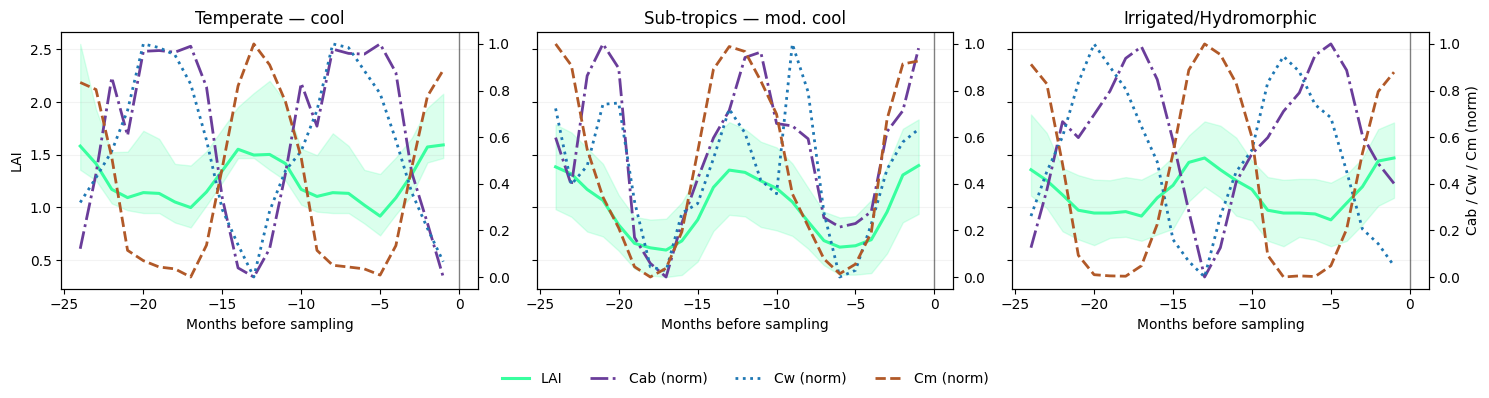}
    \caption{Trajectories of canopy traits estimated from PROSAIL inversion over 24 months before sampling, stratified by AEZ. LAI is shown with median and interquartile range; Cab, Cw, and Cm are normalized for comparability. Trait phenology differs by AEZ, reflecting contrasts in productivity, stress response, and allocation strategies that mediate soil–plant nutrient exchange.}
    \label{fig:prosail_sequences}
\end{figure}

\textbf{Radiative transfer traits (\(o,t\)).} To move beyond indices, canopy structural and biochemical traits were retrieved via an RTM inversion \cite{Jacquemoud2009PROSAIL}. Leaf area index (LAI), chlorophyll (Cab), leaf water (Cw), and dry matter (Cm) trajectories (Figure~\ref{fig:prosail_sequences}) capture physiological contrasts: stable cycles in temperate regions, sharp oscillations under subtropical climatic stress, and extended LAI in irrigated systems. These traits provide mechanistic links to vegetation physiology, stress adaptation, and nutrient assimilation.

\textbf{Representation learning (\(o,t\)).} To complement physics-based traits, we integrate embeddings from the Presto foundation model trained on Sentinel-2 EO reflectance (10 m, $\sim$5 days) \cite{tseng2024lightweightpretrainedtransformersremote}. These embeddings capture higher-order temporal dependencies, lagged responses, seasonal transitions, and nonlinear vegetation–climate interactions directly from point-referenced sequences, enabling richer representations than handcrafted indices.

\textbf{Bare-soil composites (\(p,t\)).}  
During periods of sparse vegetation cover, we derived bare-soil reflectance composites using spectral mixture analysis \cite{essd-16-1333-2024} and soil line criteria \cite{Baret1993SoilLine}. Unlike vegetation-based proxies, which only indirectly infer belowground processes, bare-soil composites provide direct optical observations of the soil surface, capturing surface moisture, mineralogical, and textural properties that correspond to parent material (\(p\)). They thus serve as a complementary window into soil conditions that are otherwise masked by canopy cover, providing direct constraints linking aboveground dynamics to belowground nutrient concentrations.

\subsection{Modeling Approach and Validation}
\label{sec:modeling_approach}

\subsubsection{Feature Selection}

The initial set of covariates comprised over $15{,}000$ variables derived from static and dynamic layers, including climate descriptors, terrain metrics, vegetation indices, radiative transfer traits, bare-soil composites, and learned embeddings. The large size of this feature space reflects the integration of multi-temporal inputs, spectral indices, physics-based canopy traits, and representation embeddings. Such high-dimensional feature spaces pose three major challenges \cite{Dormann2013Ecography}:
\begin{enumerate}
    \item Multicollinearity, which inflates variance and hinders interpretability,
    \item Instability of selected predictors across AEZs,
    \item Risk of overfitting when sample sizes are limited relative to feature count.
\end{enumerate}

\begin{figure}[!htbp]
\centering
\includegraphics[width=0.9\linewidth]{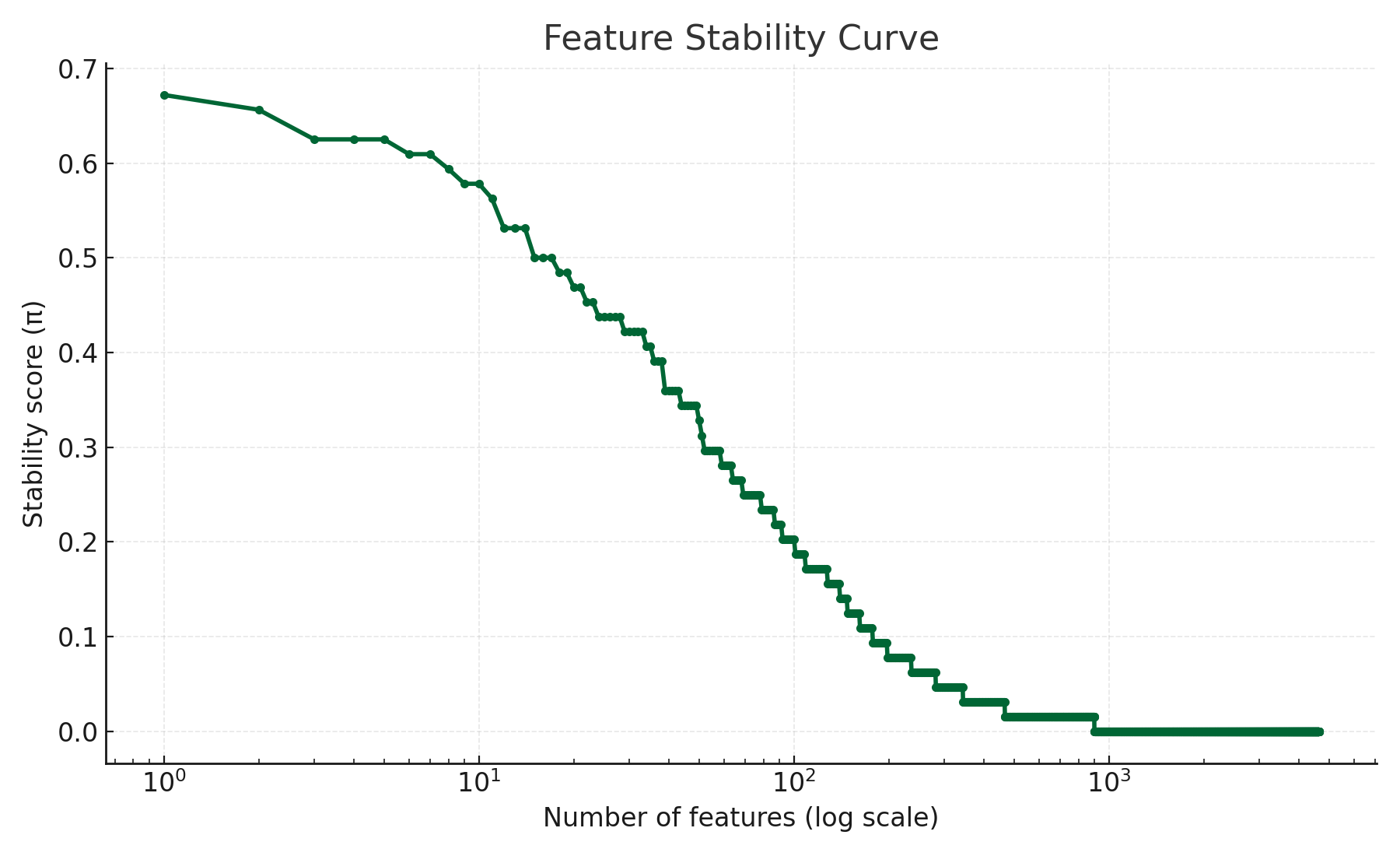}
\caption{Feature stability curve. Points show stability scores (\(\pi\)) for ranked features; x–axis shows the cumulative number of features.}
\label{fig:feat_select}
\end{figure}

To manage this high-dimensional space, we employed a two-stage reduction strategy for each targeted variable. First, we filtered out highly redundant predictors by removing variables with excessive pairwise correlation, thereby retaining a representative yet less collinear subset. Second, we applied a randomized stability selection framework \cite{Meinshausen2010JRSSB} on Extreme Gradient Boosting (XGB) models over 64 iterations, which evaluates the relevance of predictors under repeated resampling. By aggregating importance scores across iterations, this procedure identifies variables that remain consistently informative, thereby prioritizing features with robust predictive value across diverse AEZs while reducing the risk of overfitting (Table~\ref{tab:feature_reduction}). For SOC, the resulting stability curve shows a rapid decline in feature selection probabilities, with only a small subset of features being consistently stable (Figure~\ref{fig:feat_select}). This suggests that the predictive signal is concentrated in a relatively small group of variables, while the majority provides little robust contribution. This highlights the difficulty of identifying reliable covariates for estimating soil properties because nutrient-related processes operate across multiple spatial and temporal scales, producing nonlinear and weakly separable signals that make it difficult for individual covariates to consistently capture soil variation across AEZs.

\begin{table}[!htbp]
\centering
\small
\caption{Number of features per nutrient after each stage}
\label{tab:feature_reduction}
\begin{tabular}{lrrrrr}
\toprule
 & SOC & N  & P & K  & pH \\
\midrule
Initial   & 15{,}951 & 15{,}951 & 15{,}951 & 15{,}951 & 15{,}951 \\
Stage 1   & 4{,}578  &  5{,}144 & 5{,}262 & 4{,}887 & 5{,}477 \\
Stage 2   & 177      &    189   &   186   &   199   &   196 \\
\bottomrule
\end{tabular}
\end{table}

\begin{figure}[!htbp]
\centering
\includegraphics[width=0.85\linewidth]{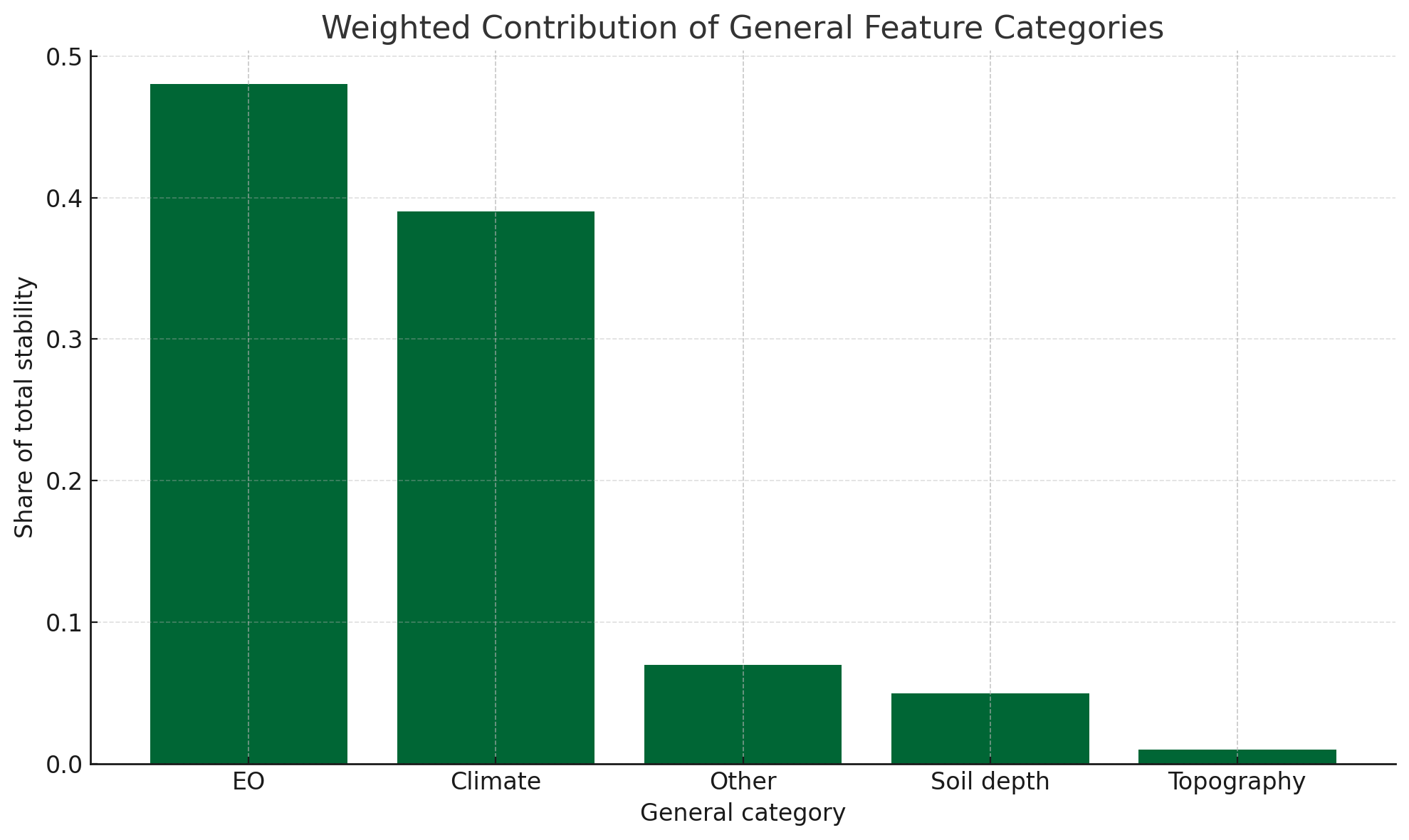}
\caption{Weighted contribution of general feature categories to overall stability. Bars represent the share of total stability contribution by each category.}
\label{fig:feat_cat}
\end{figure}

In addition to the overall stability curve, we aggregated stability contributions by broader feature categories to assess which ones provided the most consistent predictors (Figure~\ref{fig:feat_cat}). This analysis highlights the dominance of EO-derived covariates among the selected features, relative to climate, depth, and other categories.

\subsubsection{Model Training and Calibration}

Soil nutrient variables are characterized by strong skewness, interdependence, and heterogeneity across AEZs \cite{Zhang2019HeterogeneitySoil}. To capture these patterns, we trained machine learning models that integrate environmental covariates with nutrient responses while accounting for shared structure among soil properties. We used classical machine learning models, such as XGB, over deep learning models due to the interpretability and computational efficiency for tabular data. Since representative features are different across targets, each target has its own feature selection and modeling pipeline.

To reduce systematic bias, we applied calibration procedures stratified by AEZ, ensuring that adjustments respected environmental contrasts rather than enforcing a single global correction \cite{Meyer2018SpatialStats}. In addition, we implemented uncertainty quantification via conformal prediction \cite{Shafer2008JMLR}. These intervals allow us to report not only point estimates but also the reliability of predictions across heterogeneous landscapes.

\subsection{Model Evaluation Framework}
\label{sec:evaluation}

\subsubsection{Data Splitting and Statistical Evaluation}

Random cross-validation would overestimate accuracy by mixing nearby or environmentally similar samples across folds \cite{Roberts2017Ecography}. To avoid this, we combined spatial blocking, agro-ecological stratification, and distributional diagnostics.

\begin{figure}[!htb]
    \centering
    \includegraphics[width=1\textwidth]{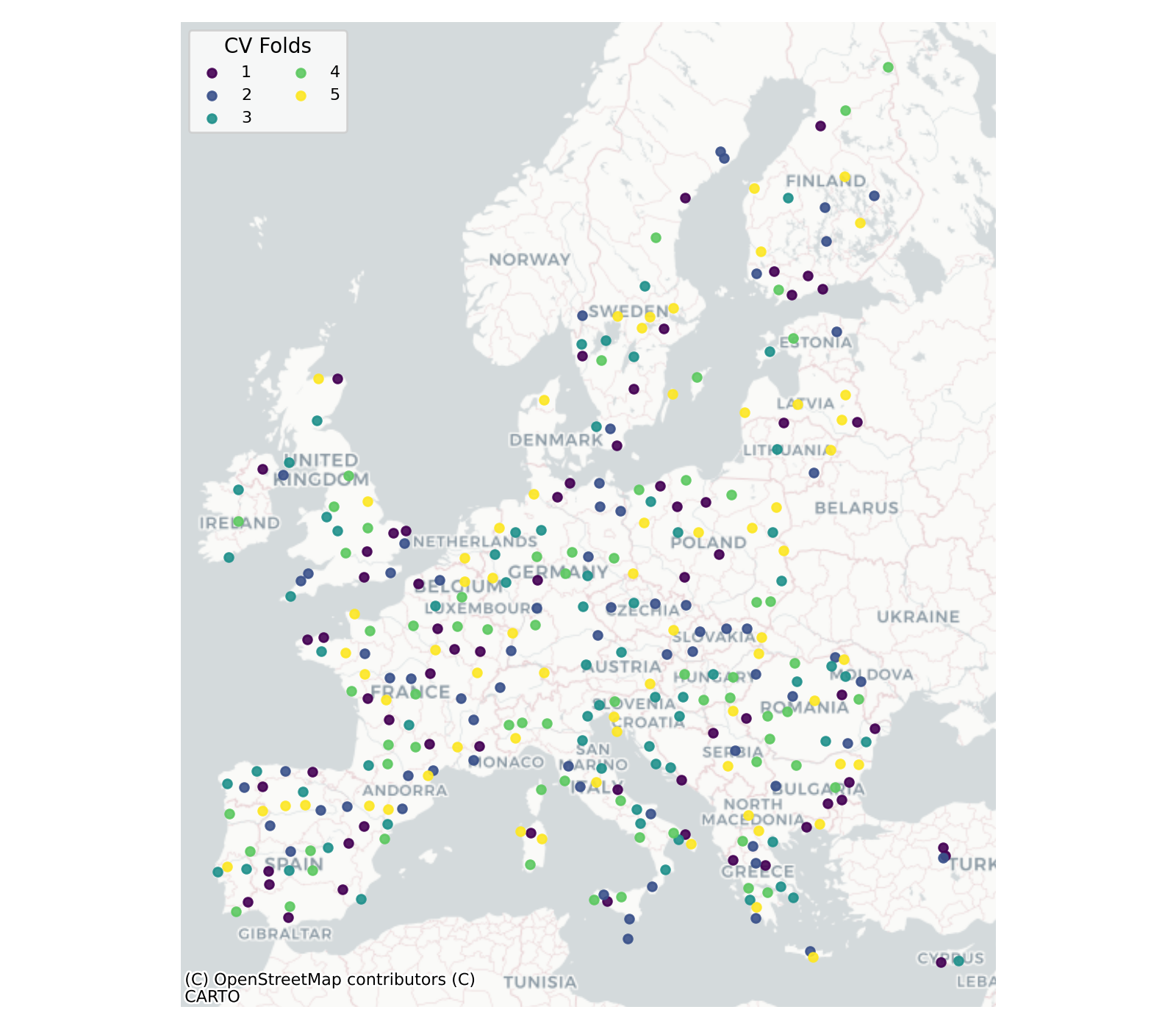}
    \caption{Cross-validation fold allocation across Europe. Sampling locations are aggregated into spatial blocks of approximately $100\,\mathrm{km}$ to mitigate spatial leakage. Distinct colors denote fold membership.}
    \label{fig:cv_folds}
\end{figure}

\paragraph{Spatial grouping.} Sampling locations were aggregated into spatial blocks of approximately $100\,\mathrm{km}$. 
These blocks were treated as indivisible units for partitioning: all samples within a block were assigned to the same fold, ensuring that no two geographically proximate points contributed to training and testing. 
This design mitigates short-range spatial autocorrelation and enforces evaluation on genuinely independent regions (Figure~\ref{fig:cv_folds}).

\begin{figure}[!htbp]
    \centering
    \includegraphics[width=0.8\textwidth]{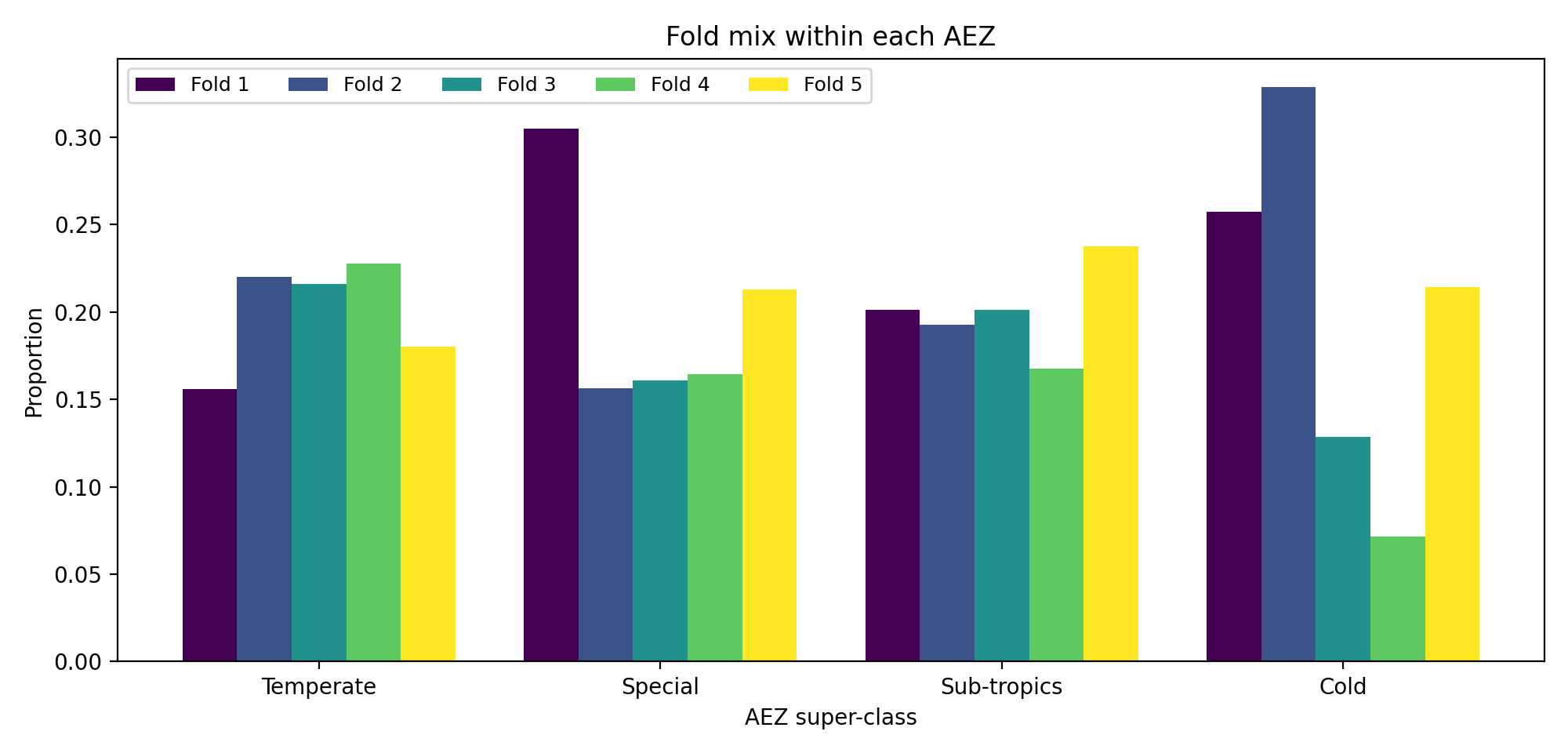}
    \caption{Distribution of samples across aggregated individual AEZ classes into four super-classes (Subtropics, Temperate, Cold, and Special Conditions, such as irrigated systems or steep terrain). Stratification ensures that minority AEZs remain represented.}
    \label{fig:aez_balance}
\end{figure}

\textbf{Agro-ecological stratification.} Because AEZs vary in both sample density and soil properties, partitioning was stratified by AEZ. Each fold contained an approximately proportional representation of samples from all AEZs, preventing dominance by temperate–cool systems and preserving minority environments such as terrain-limited or subtropical zones (Figure~\ref{fig:aez_balance}).

\textbf{Distributional diagnostics.} We confirmed that folds differed significantly, providing a more realistic evaluation of model generalization. The Kolmogorov--Smirnov (KS) test \cite{Kolmogorov1933} and Anderson–Darling (AD) test \cite{AndersonDarling1952} showed distributional differences in soil properties (Table~\ref{tab:ks_ad}), while the nearest neighbor analysis confirmed that the test blocks were geographically separated from training blocks (Table~\ref{tab:nn_dist}).

\begin{table}[!htb]
    \centering
    \caption{KS and AD tests comparing train vs.\ test distributions of SOC. Significant \(p\)-values indicate that the test partitions represent distinct distributions.}
    \label{tab:ks_ad}
    \begin{tabular}{lcc}
        \toprule
        \textbf{Fold} & \textbf{KS \(p\)-value} & \textbf{AD \(p\)-value} \\
        \midrule
        1 & 0.002 & 0.001 \\
        2 & 0.014 & 0.008 \\
        3 & 0.021 & 0.012 \\
        4 & 0.005 & 0.003 \\
        5 & 0.018 & 0.010 \\
        \bottomrule
    \end{tabular}
\end{table}

\begin{table}[!htb]
    \centering
    \caption{Nearest-neighbor distances (km) between test blocks and the closest training block. Large median values confirm the spatial independence of the evaluation sets.}
    \label{tab:nn_dist}
    \begin{tabular}{lccc}
        \toprule
        \textbf{Fold} & \textbf{Mean (km)} & \textbf{Median (km)} & \textbf{95th Percentile (km)} \\
        \midrule
        1 & 121.3 & 108.5 & 255.7 \\
        2 & 115.8 & 104.2 & 240.1 \\
        3 & 123.5 & 111.0 & 262.4 \\
        4 & 118.6 & 106.7 & 247.5 \\
        5 & 120.2 & 107.9 & 251.9 \\
        \bottomrule
    \end{tabular}
\end{table}

\textbf{Evaluation metrics.}  
Performance was quantified across four complementary dimensions.  
First, accuracy was measured by root mean square error (RMSE) and mean absolute error (MAE),
\[
\mathrm{RMSE}=\sqrt{\tfrac{1}{n}\sum (y_i-\hat{y}_i)^2},\quad
\mathrm{MAE}=\tfrac{1}{n}\sum |y_i-\hat{y}_i|.
\]
Second, agreement was assessed using the concordance correlation coefficient (CCC),
\[
\mathrm{CCC} = 
\frac{2\rho\,\sigma_y\sigma_{\hat{y}}}
{\sigma_y^2 + \sigma_{\hat{y}}^2 + (\mu_y - \mu_{\hat{y}})^2}
\]%
\cite{Lin1989CCC}

and Willmott’s index of agreement,
\[
d_{1.5} =
1 - \frac{\sum |y_i - \hat{y}_i|^{1.5}}
         {\sum \bigl(|\hat{y}_i - \bar{y}| + |y_i - \bar{y}|\bigr)^{1.5}}
\]%
\cite{Willmott2012d}

Third, we quantified distributional fidelity via the ratio of performance to interquartile distance (RPIQ),
\[
\mathrm{RPIQ} = \frac{\mathrm{IQR}(y)}{\mathrm{RMSE}}
\]%
\cite{rs12071095}
with log-transformed targets to reduce skew.  
Fourth, we measured bias and reliability using bias and normalized root mean square error (NRMSE$_{\min\text{--}\max}$).  

Finally, for uncertainty quantification, we applied conformal prediction to calibrate intervals. With calibration residuals \( r_i = |y_i - \hat{y}_i| \), the \( (1-\alpha) \) prediction interval for each point is
\[
[L_i, U_i] = [\hat{y}_i - q_{1-\alpha}, \hat{y}_i + q_{1-\alpha}],
\]
where \( q_{1-\alpha} \) is the empirical \( (1-\alpha) \)-quantile of the calibration residuals \( \{ r_i \} \) \cite{Angelopoulos2021}.

We evaluated the intervals on their coverage and sharpness. The predictive interval coverage probability (PICP) measures the proportion of true values falling within the predicted intervals, which should ideally be close to the nominal level \( (1-\alpha) \):
\[
\text{PICP} = \frac{1}{n} \sum_{i=1}^{n} \mathbb{1}\{ y_i \in [L_i, U_i] \}.
\]

For sharpness, which reflects the usefulness of the intervals, we report two metrics. The primary measure is the mean prediction interval width (MPIW):
\[
\text{MPIW} = \frac{1}{n} \sum_{i=1}^{n} (U_i - L_i).
\]

To provide a scale-free measure of sharpness, we also compute the prediction interval normalized average width (PINAW). This metric normalizes the MPIW by the range of the target variable, \( R = y_{\max} - y_{\min} \):
\[
\text{PINAW} = \frac{\text{MPIW}}{R}.
\]

All metrics were computed per fold and for the test set, being further disaggregated by AEZ.

\section{Results}
\label{sec:results}

\subsection{Overall Predictive Performance}
We report performance under two complementary scenarios: 
(i) out-of-fold (OOF) cross-validation estimates, which represent a conservative assessment under spatially disjoint splits (Table~\ref{tab:oof_perf}), and 
(ii) independent test set evaluation, reflecting an optimistic upper bound on performance when models are trained on the full calibration data (Table~\ref{tab:test_perf}). 

\begin{table}[!htbp]
\centering
\caption{OOF cross-validation performance (pessimistic scenario). Values averaged across folds; $n$ sums labeled samples across folds.}
\label{tab:oof_perf}
\begin{tabular}{lcccccccc}
\toprule
Target & RMSE & MAE & CCC$_{\log_{1p}}$ & $d_p^{1.5}$ & RPIQ & Bias & NRMSE$_{\min\text{--}\max}$ & $n$ \\
\midrule
SOC & 7.92 & 5.34 & 0.75 & 0.71 & 1.51 & -1.27 & 0.13 & 18{,}027 \\
N & 0.59 & 0.44 & 0.73 & 0.73 & 1.85 & -0.06 & 0.14 & 16{,}000 \\
P & 23.24 & 15.62 & 0.45 & 0.55 & 1.33 & -6.51 & 0.15 & 12{,}097 \\
K & 114.40& 83.90 & 0.38 & 0.53 & 1.42 & -26.76& 0.19 & 13{,}109 \\
pH & 0.72 & 0.55 & 0.67 & 0.74 & 2.36 & -0.02 & 0.13 & 16{,}884 \\
\bottomrule
\end{tabular}
\end{table}

\begin{table}[!htbp]
\centering
\caption{Independent test set performance. Models were trained on the full calibration data and evaluated on held-out test samples.}
\label{tab:test_perf}
\begin{tabular}{lcccccccc}
\toprule
Target & RMSE & MAE & CCC$_{\log_{1p}}$ & $d_p^{1.5}$ & RPIQ & Bias & NRMSE$_{\min\text{--}\max}$ & $n$ \\
\midrule
SOC & 7.81 & 5.12 & 0.77 & 0.72 & 1.47 & -1.23 & 0.13 & 6{,}153 \\
N & 0.56 & 0.40 & 0.77 & 0.76 & 1.79 & -0.03 & 0.13 & 5{,}783 \\
P & 22.00 & 14.96 & 0.53 & 0.60 & 1.43 & -6.67 & 0.14 & 4{,}090 \\
K & 107.50& 76.99 & 0.58 & 0.63 & 1.58 & -25.64 & 0.17 & 4{,}274 \\
pH & 0.74 & 0.57 & 0.67 & 0.75 & 2.30 & +0.03 & 0.10 & 6{,}076 \\
\bottomrule
\end{tabular}
\end{table}

\subsection{Error Structure}
\begin{figure}[!htbp]
\centering
\includegraphics[width=1\textwidth]{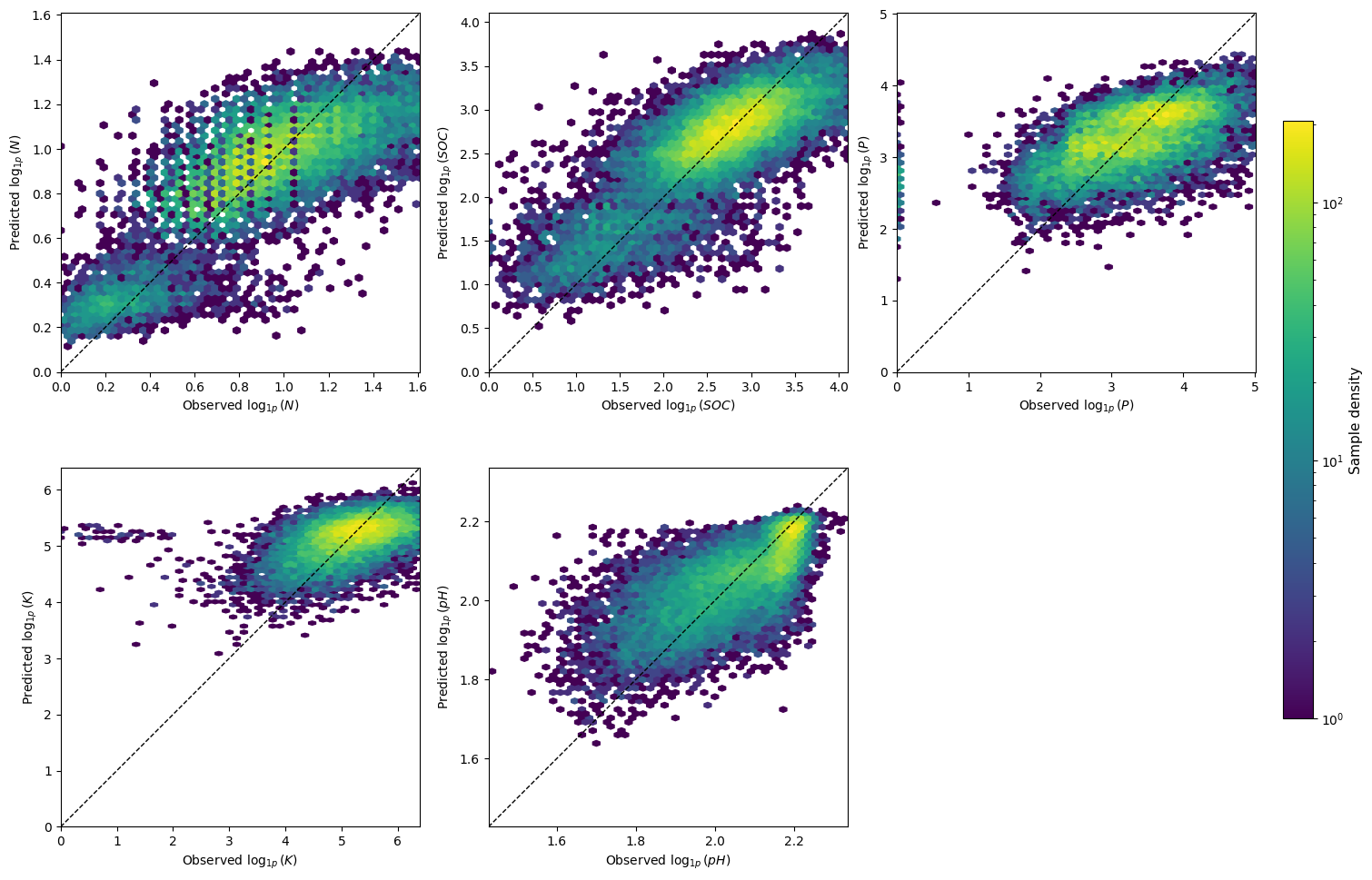}
\caption{Observed vs.\ predicted values for SOC, N, P, K, and pH on the $\log_{1p}$ scale.}
\label{fig:obs_pred}
\end{figure}

Figure~\ref{fig:obs_pred} shows observed–predicted scatterplots for all targets 
(SOC, N, P, K, and pH) on the $\log_{1p}$ scale. 
There is an underestimation at the high end and a slight overestimation at the low end for all targets. This supports the use of the $\log_{1p}$ transformation, which reduces skewness and heteroscedasticity and preserves relative ranking. SOC, N, and pH have the strongest alignment between the observed and predicted values, while P and K achieve moderate alignment, which is expected with the higher variability and the weaker correlation with the covariates. From the density plots, the predicted values closely follow the observed values in the ranges where most samples occur.

\subsection{Performance Across AEZs}
\begin{figure}[!htbp]
\centering
\includegraphics[width=0.95\textwidth]{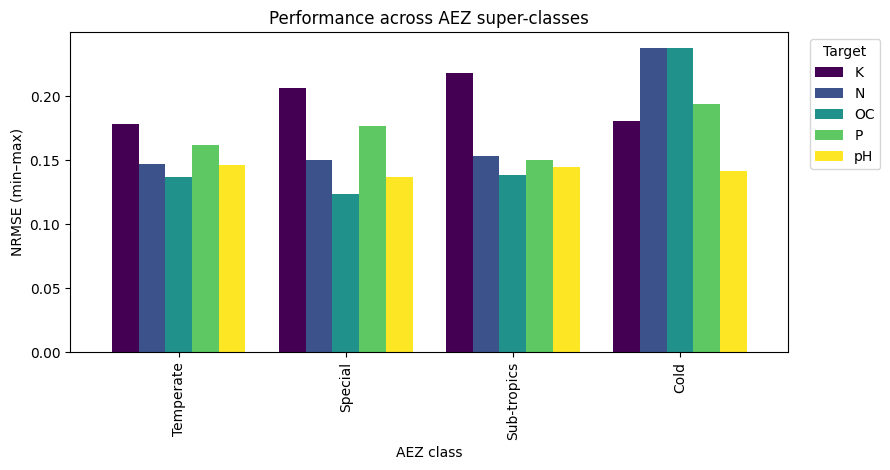}
\caption{Performance across AEZ super-classes. Bars show min--max normalized RMSE (NRMSE$_{\min\text{--}\max}$) for each nutrient target, aggregated as weighted means by sample count.}
\label{fig:aez_superclass_perf}
\end{figure}

Predictive performance remains broadly stable across the main AEZs. 
We aggregated individual AEZ classes into four super-classes (Subtropics, Temperate, Cold, and Special Conditions, such as irrigated or steep terrain) to provide a compact overview. 
Figure~\ref{fig:aez_superclass_perf} shows NRMSE (min--max normalized) for each nutrient target across these super-classes. 
Performance is mostly consistent across zones, with modest degradation in cold and terrain-limited environments where sampling density is lower or environmental variability is higher.

\subsection{Performance Across Soil Depths}
\begin{figure}[!htbp]
\centering
\includegraphics[width=0.9\textwidth]{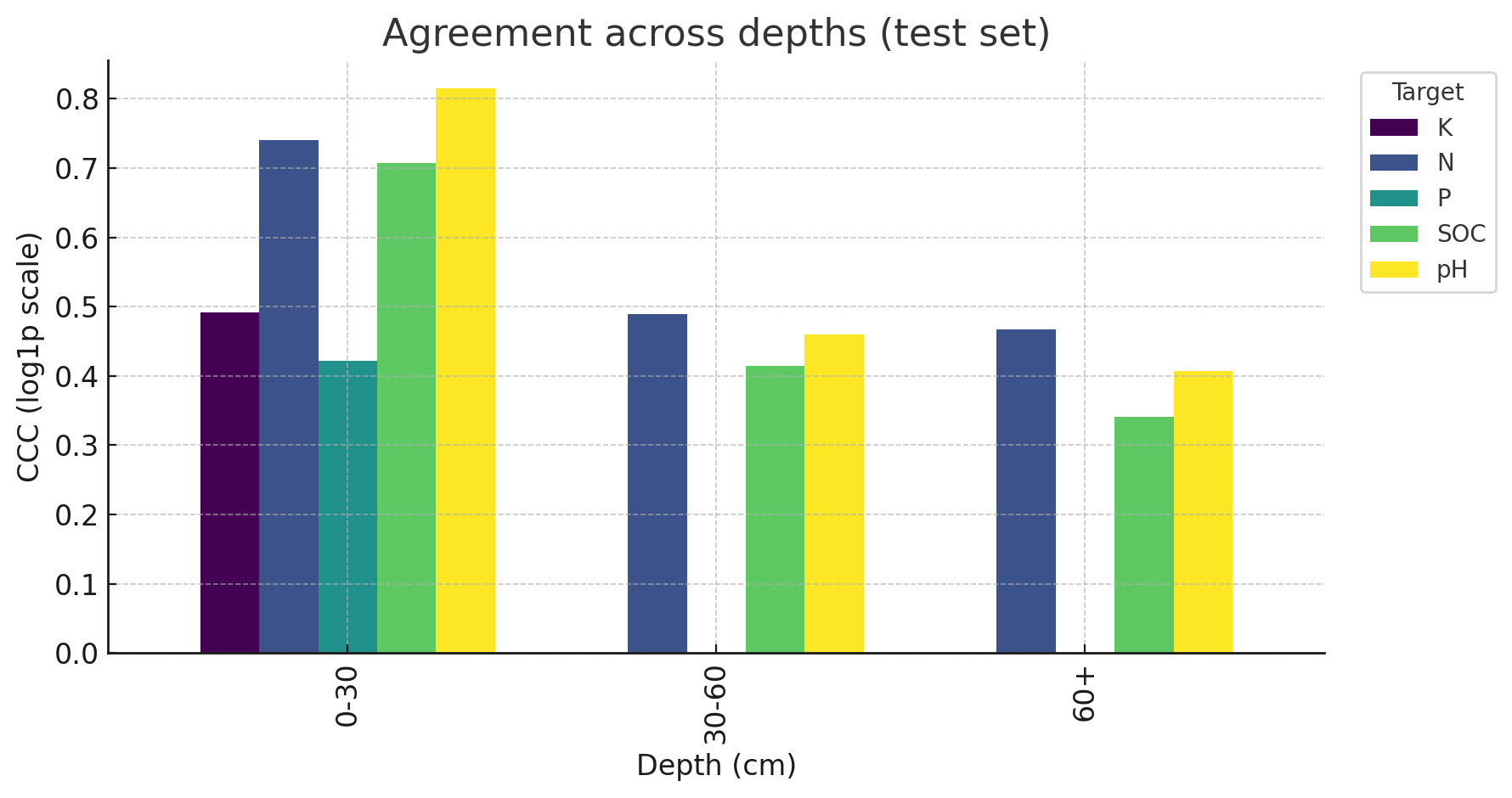}
\caption{Concordance (CCC, \(\log_{1p}\) scale) across depth categories on the independent test set.}
\label{fig:depth_perf_ccc}
\end{figure}

Predictive performance shows a clear dependence on sampling depth. 
Figure~\ref{fig:depth_perf_ccc} summarizes CCC across the three main depth categories (0--30, 30--60, and 60+ cm). For all targets, models achieve their strongest performance in the 0--30 cm layer, which is also the most densely sampled and agronomically relevant horizon. At deeper horizons, performance systematically decreases. This pattern reflects both the limited physical penetration of surface reflectance signals and the reduced number of samples available below 30 cm. K and P could only be robustly assessed in the 0--30 cm range; test sets at greater depths contained fewer than ten samples, yielding unstable and uninterpretable metrics. These results underscore that current remote sensing-based approaches are better suited for topsoils. While the models retain some predictive signal for SOC, N, and pH at depth, reliable mapping beyond the plough layer will likely require more extensive ground sampling or complementary sensing modalities.

\subsection{Uncertainty Calibration}
We assessed the quality of predictive uncertainty using conformal prediction intervals at the 90\% nominal level. Across OOF predictions, all nutrients achieved the nominal coverage of 90\% (Table~\ref{tab:unc_global}), indicating well-calibrated intervals. 
Interval widths reflect the scale of each nutrient: SOC intervals averaged $\sim$25~g/kg, N intervals $\sim$1.8~g/kg, and pH intervals $\sim$2.5~pH units. 
Normalized widths (PINAW $\approx 0.40$–0.45) suggest that intervals occupy less than half of the observed range, providing reasonably sharp uncertainty bounds. 

Stratification by AEZs for SOC (Table~\ref{tab:unc_aez}) shows that coverage remains stable around the nominal 90\% across diverse environments. 
Interval widths vary with AEZ, with broader intervals observed in terrain-limited or subtropical regions (e.g., AEZ 21, 26, 15), consistent with higher environmental heterogeneity and reduced sampling density. 

\begin{table}[!htbp]
\centering
\caption{Uncertainty metrics from OOF predictions. PICP: prediction interval coverage probability, MPIW: mean prediction interval width, PINAW: normalized average width}
\label{tab:unc_global}
\begin{tabular}{lccccc}
\toprule
Nutrient & $n$ & PICP & MPIW & PINAW \\
\midrule
SOC & 17{,}659 & 0.90 & 24.65 & 0.41\\
N   & 15{,}653 & 0.90 & 1.79 & 0.44\\
pH  & 16{,}511 & 0.90 & 2.45  & 0.39\\
\bottomrule
\end{tabular}
\end{table}

\begin{table}[!htbp] 
\centering 
\caption{Uncertainty metrics for SOC, stratified by AEZs under OOF evaluation. Rankings are based on MPIW and PINAW (lower is better).} 
\label{tab:unc_aez} 
\begin{tabular}{lccccc} 
\toprule 
AEZ & $n$ & PICP & MPIW & PINAW & Avg. Rank \\ 
\midrule 
16 (Temperate, moderate; dry) & 387 & 0.90 & 19.70 & 0.32 & 1.0 \\ 
13 (Subtropics, cool; semi-arid) & 751 & 0.90 & 21.53 & 0.36 & 2.0 \\ 
27 (Irrigated soils) & 2{,}855 & 0.90 & 22.37 & 0.37 & 3.0 \\ 
20 (Temperate, cool; moist) & 6{,}352 & 0.90 & 23.31 & 0.39 & 4.0 \\ 
11 (Subtropics, m.cool; sub-humid) & 1{,}341 & 0.90 & 23.57 & 0.39 & 4.5 \\ 
28 (Hydromorphic soils) & 850 & 0.90 & 23.82 & 0.39 & 5.0 \\ 
17 (Temperate, moderate; moist) & 260 & 0.90 & 24.54 & 0.41 & 7.0 \\ 
14 (Subtropics, cool; sub-humid) & 1{,}138 & 0.90 & 25.96 & 0.43 & 8.0 \\ 
21 (Temperate, cool; wet) & 2{,}939 & 0.90 & 28.42 & 0.47 & 9.0 \\ 
12 (Subtropics, m.cool; humid) & 146 & 0.90 & 33.50 & 0.56 & 10.0 \\ 
26 (Severe terrain limits) & 402 & 0.90 & 35.59 & 0.59 & 11.0 \\ 
15 (Subtropics, cool; humid) & 238 & 0.90 & 38.09 & 0.63 & 12.0 \\ 
\bottomrule 
\end{tabular} 
\end{table}
\section{Discussion}
\label{sec:discussion}

\subsection{Spatial Heterogeneity and Imbalance Effect}
\label{sec:spatial_imbalance}

Models risk overfitting to over-sampled AEZs and may fail to generalize to underrepresented AEZs, which also exhibit higher environmental variability (e.g., Subtropics, Severe terrain limits). To address this imbalance and ensure robust estimates across various zones, our stratified sampling and spatial blocking have mitigated most of the spatial autocorrelation, resulting in reduced disparities in the predictive errors across AEZs. This highlights the model's ability to reflect spatial heterogeneity under a strict spatially blocked approach. There is potential to improve further with more targeted data from specific AEZs or through model fine-tuning, increasing its operational utility.

\subsection{Hybrid Modeling Strategy}

The hybrid modeling approach integrates indirect proxies of soil processes with direct reflectance signals. By combining the philosophies of modeling soil nutrients, we leverage the strengths of both indirect and direct modeling methods to offer a novel perspective on soil nutrient analysis. While initiatives like AI4SoilHealth \cite{Minarik2024SoilHealthDataCube} focus on digital soil mapping using indirect proxies, they do not account for direct reflectance from bare-soil, which provides a measure of soil reflectance. Conversely, World-Soils \cite{VANWESEMAEL2024117113} mainly emphasizes bare-soil for croplands, omitting aboveground dynamics and static covariates. Our approach builds on the previous methods by combining physics-informed features through RTM with representations learned via foundation models, offering a balance between predictive performance and interpretability. The RTM-based features provide direct, physically informed insights into soil properties, while the foundation models capture complex, non-linear relationships in the data.

\subsection{Importance of Satellite Data}

The study has shown that EO-derived covariates contribute the most to the predictive signal, making them central to the modeling of soil properties at the continental scale. This is important considering the requirement of a scalable yet robust system in which satellites enable operational-scale mapping of soil properties. Table~\ref{tab:satellite_breakdown} shows that bare-soil composites and derived features from vegetation time-series account for more of the stability share. Radiative transfer and foundation-model features also contribute, though their stability contribution remains limited. The addition of hyperspectral imaging is expected to show an increase in performance, especially targeted at bare-soil observation strategies. Upcoming missions such as  ESA’s Copernicus Hyperspectral Imaging Mission for the Environment (CHIME), with continuous hyperspectral coverage at 30m resolution, will enable operational applications at scale \cite{Nieke2023CHIME, ESA-GoingHyperspectral}. 

\begin{table}[!htbp]
\centering
\caption{Breakdown of stability contribution within the EO category. Values show total stability and relative share.}
\label{tab:satellite_breakdown}
\begin{tabular}{lcc}
\toprule
Sub-category       & Total stability & Share \\
\midrule
Bare-soil composites  & 10.27           & 45.8\% \\
Vegetation and productivity metrics  & 9.16            & 40.9\% \\
Radiative transfer traits   & 1.72            & 7.7\%  \\
Representation learning  & 1.25            & 5.6\%  \\
\bottomrule
\end{tabular}
\end{table}

\subsection{Model Performance \& Comparison}

The model’s predictive performance was assessed using both cross-validation and an independent test set, under strict spatially disjoint splits. Results indicate robust performance across all nutrients for regional monitoring and trend analysis. For SOC, N, and pH, the error levels under strict spatially disjoint validation sets (compared to soil sample uncertainty) suggest that the models are also suitable for agronomic applications. This is reflected in the error structure, which shows strong correlation for SOC and N in log-space, with degradation in the original space, showing that models struggle to capture extreme values. This is primarily due to limited data availability in AEZs that present such variability, although such outliers can be managed in operational settings through additional sampling. In contrast, the predictions for P and K require additional data or regional fine-tuning to achieve accuracy sufficient for precise agronomic decision making. This highlights the difficulty in assessing P and K due to the lack of direct spectral response of such nutrients, compared to SOC and N, whose spectral responses are more visible with environmental covariates. When compared with recent continental-scale studies that leverage LUCAS-derived SOC and nutrient maps, such as AI4SoilHealth \cite{Minarik2024SoilHealthDataCube} and World-Soils \cite{VANWESEMAEL2024117113}, our models produce predictions that match or exceed published performance under spatially strict, pessimistic validation (Table~\ref{tab:ai4soilhealth}, Table~\ref{tab:soc_worldsoils}).

\begin{table}[!htbp]
\centering
\caption{Comparison of nutrient prediction performance between this study and AI4SoilHealth. Best values per nutrient and metric are in bold.}
\label{tab:ai4soilhealth}
\begin{tabular}{lccc}
\toprule
& \multicolumn{2}{c}{CCC$_{\log_{1p}}$} \\
\cmidrule{2-3} 
Nutrient & This study & AI4SoilHealth \\
\midrule
SOC & \textbf{0.77} & 0.58 \\
N & \textbf{0.74} & 0.72   \\
pH & 0.67 & \textbf{0.73}  \\
P & \textbf{0.53} & 0.30   \\
K & 0.58 & \textbf{0.72}  \\
\bottomrule
\end{tabular}
\end{table}

\begin{table}[!htbp]
\centering
\caption{Comparison of SOC prediction performance between this study and World-Soils. Best values per metric are in bold.}
\label{tab:soc_worldsoils}
\begin{tabular}{lcc}
\toprule
Study & RMSE (g/kg) & RPIQ \\
\midrule
This study  & \textbf{7.8}  & \textbf{1.47} \\
World-Soils (Bare-Soil) & 18.07  & 0.73 \\
\bottomrule
\end{tabular}
\end{table}

 These papers highlight the two distinct modeling philosophies when it comes to remote soil nutrient analysis, meaning direct measurement through bare-soil reflectance (World-Soils) and indirect measurement through proxies (AI4SoilHealth). It is also important to interpret these comparisons in light of methodological differences: for instance, AI4SoilHealth and World-Soils include all land covers, whereas our study focuses on cropland only, and our evaluation framework enforces AEZ stratification and statistically distinct sets. These design choices contribute to more conservative but robust performance estimates. Overall, the findings position our modeling and evaluation framework as a robust and competitive approach for large-scale soil nutrient estimation, particularly under stringent spatial validation.

\section{Conclusion}
\label{sec:conclusion}
This study addresses methodological limitations in two areas: 
(i) developing a scalable, data-driven framework that supports environmentally oriented and economically viable agricultural practices while reducing reliance on labor-intensive sampling, and
(ii) refining remote sensing approaches for soil property estimation together with evaluation protocols tailored to stratified, spatially blocked, and statistically distinct validation sets.
We conduct the study across European croplands, utilizing a harmonized dataset standardized to the LUCAS topsoil studies. The hybrid modeling approach combined direct and indirect modeling frameworks, while extending them with physics-informed features via RTM inversion and spatio-temporal embeddings from foundation models. The robustness was demonstrated under spatially stratified cross-validation and independent testing across AEZs, yielding for SOC a MAE of 5.12 g/kg with CCC of 0.77, for N an MAE of 0.44 g/kg and CCC of 0.77, for P an MAE of 14.96 mg/kg and CCC of 0.53, for K an MAE of 76.99 mg/kg and CCC of 0.58, and for pH an MAE of 0.55 and CCC of 0.67. Disaggregating performance by AEZ shows that the models perform consistently across most zones, with degradation only in under-represented or environmentally challenging AEZs. This highlights the benefits of stratified and spatially blocked evaluation. These results show that the SOC, N, and pH models are reliable for both regional trend analysis and potentially parcel-level agronomic decisions. At the same time, P and K require either more data in the underrepresented AEZs or regional fine-tuning to become reliable for agronomic decisions. With conformal prediction achieving calibrated 90\% uncertainty coverage, the framework demonstrates uncertainty reliability for decision making. This provides a scalable basis for digital soil mapping. Future work will focus on expanding the dataset for global applicability, including hyperspectral data, and testing the feasibility of practical usage in nutrient management for agriculture.

\section*{Declaration of competing interest}
David Șeu and Erik Maidik are co-founders and equity holders of CO2 Angels SRL. Călin Andrei and Gabriel Cioltea work at CO2 Angels. 
The methods reported here will be implemented in CO2 Angels’ products and may directly benefit the company. The remaining authors declare no competing interests.

\section*{Acknowledgements}
Special thanks are extended to the EuroCC Netherlands program for providing access to high-performance computing resources. The authors also acknowledge the contributions of Călin Mihai Cuc (Fabrica de Compost) for expertise in agricultural science, and Emil Ellekær Hjortflod (Penneo) for insights into agricultural financial aspects. Finally, the authors express their appreciation to colleagues and collaborators at CO2 Angels and the European Space Agency \(\Phi\)-Lab for their guidance and constructive feedback throughout the study.

\section*{Data availability}
The authors do not have permission to share data.

\section*{Acronyms and Abbreviations}
\begin{longtable}{ll}
\caption{List of acronyms and abbreviations.} \\
\toprule
\textbf{Acronym} & \textbf{Meaning} \\
\midrule
\endfirsthead

\multicolumn{2}{c}%
{\tablename\ \thetable\ -- \textit{continued from previous page}} \\
\toprule
\textbf{Acronym} & \textbf{Meaning} \\
\midrule
\endhead

\midrule \multicolumn{2}{r}{\textit{Continued on next page}} \\
\endfoot

\bottomrule
\endlastfoot

3D+T & Three-dimensional plus time \\
AD & Anderson--Darling (test) \\
AEZ & Agro-Ecological Zone \\
Cab & Leaf chlorophyll content \\
CERES & Clouds and the Earth’s Radiant Energy System \\
CHIME & Copernicus Hyperspectral Imaging Mission for the Environment \\
CHELSA & Climatologies at High Resolution for the Earth’s Land Surface Areas \\
CCC & Concordance Correlation Coefficient \\
Cm & Leaf dry matter content \\
Cw & Leaf water content \\
EO & Earth Observation \\
ERA5 & ECMWF Reanalysis v5 \\
ESA & European Space Agency \\
EU & European Union \\
GAEZ & Global Agro-Ecological Zones \\
GLAD & Global Land Analysis and Discovery \\
GVFCP & Global Vegetation Fractional Cover Product \\
HLS & Harmonized Landsat--Sentinel (surface reflectance) \\
IQR & Interquartile Range \\
KS & Kolmogorov--Smirnov (test) \\
K & Exchangeable potassium \\
LAI & Leaf Area Index \\
LST & Land Surface Temperature \\
LUCAS & Land Use and Land Cover Survey (EU topsoil program) \\
MAE & Mean Absolute Error \\
MODIS & Moderate Resolution Imaging Spectroradiometer \\
MPIW & Mean Prediction Interval Width \\
N & Total nitrogen \\
NDVI & Normalized Difference Vegetation Index \\
NDTI & Normalized Difference Tillage Index \\
NRMSE$_{\min\text{--}\max}$ & Min--max normalized RMSE \\
OOF & Out-of-fold (cross-validation) \\
P & Available phosphorus \\
PICP & Prediction Interval Coverage Probability \\
PINAW & Prediction Interval Normalized Average Width \\
pH$_{\mathrm{H_2O}}$ & Soil pH in water \\
PROSAIL & PROSPECT + SAIL radiative transfer model \\
RPIQ & Ratio of Performance to Interquartile distance \\
RMSE & Root Mean Square Error \\
RTM & Radiative Transfer Model \\
SCORPAN & Soil state-factor model: $S=f(c,o,r,p,a,n,t)$ \\
SOC & Soil Organic Carbon \\
TP & Total Precipitation \\
$\,d_{1.5}$ & Willmott’s index of agreement (power 1.5) \\
XGB & Extreme Gradient Boosting \\
\end{longtable}

\bibliographystyle{unsrtnat}
\bibliography{references}

@article{Kopittke2019EnvInt,
  author  = {Kopittke, Peter M. and Menzies, Neal W. and Wang, Peng and McKenna, Bruce A. and Lombi, Enzo},
  title   = {Soil and the intensification of agriculture for global food security},
  journal = {Environment International},
  year    = {2019},
  volume  = {132},
  pages   = {105078},
  doi     = {10.1016/j.envint.2019.105078},
  url     = {https://doi.org/10.1016/j.envint.2019.105078}
}

@article{Choruma2024JAFR,
  author  = {Choruma, D. and others},
  title   = {Digitalisation in agriculture: A scoping review},
  journal = {Journal of Agriculture and Food Research},
  year    = {2024},
  volume  = {18},
  pages   = {101286},
  doi     = {10.1016/j.jafr.2024.101286},
  url     = {https://doi.org/10.1016/j.jafr.2024.101286}
}

@article{Orgiazzi2018EJSS,
  author  = {Orgiazzi, Alberto and Ballabio, Cristiano and Panagos, Panos and Jones, Arwyn and Fernandez-Ugalde, Olga},
  title   = {LUCAS Soil, the largest expandable soil dataset for Europe: a review},
  journal = {European Journal of Soil Science},
  year    = {2018},
  volume  = {69},
  number  = {1},
  pages   = {140--153},
  doi     = {10.1111/ejss.12499},
  url     = {https://doi.org/10.1111/ejss.12499}
}

@article{Norouzi2021,
  author = {Norouzi, Sarem and Sadeghi, Morteza and Liaghat, Abdolmajid and Tuller, Markus and Jones, Scott B.},
  title = {Information depth of {NIR/SWIR} soil reflectance spectroscopy},
  journal = {Remote Sensing of Environment},
  volume = {256},
  pages = {112315},
  year = {2021},
  doi = {10.1016/j.rse.2021.112315}
}

@techreport{Jones2020LUCAS2015,
  author      = {Jones, Arwyn and Fernandez-Ugalde, Olga and Scarpa, Simona},
  title       = {LUCAS 2015 Topsoil Survey: Presentation of dataset and results},
  institution = {Publications Office of the European Union},
  address     = {Luxembourg},
  year        = {2020},
  doi         = {10.2760/616084},
  url         = {https://data.europa.eu/doi/10.2760/616084}
}

@misc{JRC2020LUCAS2018,
  author       = {{Joint Research Centre (JRC)}},
  title        = {LUCAS 2018 Topsoil Data},
  year         = {2020},
  howpublished = {\url{https://esdac.jrc.ec.europa.eu/content/lucas-2018-topsoil-data}},
  note         = {European Soil Data Centre (ESDAC), Joint Research Centre}
}

@article{OConnell2022Geodrs,
  author  = {O'Connell, C. and others},
  title   = {Why soil testing is not enough: A mixed methods study of innovation adoption in agriculture},
  journal = {Journal of Environmental Management},
  year    = {2022},
  volume  = {314},
  pages   = {115027},
  doi     = {10.1016/j.jenvman.2022.115027},
  url     = {https://doi.org/10.1016/j.jenvman.2022.115027}
}

@article{vanDijk2021NatFood,
  author    = {van Dijk, M. and Morley, T. and Rau, M. L. and Saghai, Y.},
  title     = {A meta-analysis of projected global food demand and population at risk of hunger for the period 2010–2050},
  journal   = {Nature Food},
  year      = {2021},
  volume    = {2},
  pages     = {494--501},
  doi       = {10.1038/s43016-021-00322-9}
}

@misc{FAO2023SOFI,
  title     = {The State of Food Security and Nutrition in the World 2023},
  author     = {{FAO, IFAD, UNICEF, WFP and WHO}},
  year      = {2023},
  howpublished = {\url{https://doi.org/10.4060/cc3017en}}
}

@article{Ray2013PLOSONE,
  author    = {Ray, D. K. and Mueller, N. D. and West, P. C. and Foley, J. A.},
  title     = {Yield Trends Are Insufficient to Double Global Crop Production by 2050},
  journal   = {PLOS ONE},
  year      = {2013},
  volume    = {8},
  number    = {6},
  pages     = {e66428},
  doi       = {10.1371/journal.pone.0066428}
}

@article{Chowdhury2024DiB,
  author  = {Chowdhury, S. and others},
  title   = {Drivers of soil health across European Union – Data from the literature review},
  journal = {Data in Brief},
  year    = {2024},
  volume  = {54},
  pages   = {111064},
  doi     = {10.1016/j.dib.2024.111064},
  url     = {https://doi.org/10.1016/j.dib.2024.111064}
}

@article{Panagos2025GlobalChallenges,
   author  = {Panagos, P. and others},
  title   = {A Soil Monitoring Law for Europe},
  journal = {Global Challenges},
  year    = {2025},
  volume  = {9},
  number  = {3},
  pages   = {202400336},
  doi     = {10.1002/gch2.202400336},
  url     = {https://doi.org/10.1002/gch2.202400336}
}

@article{Lobell2011Science,
  author    = {Lobell, D. B. and Schlenker, W. and Costa-Roberts, J.},
  title     = {Climate Trends and Global Crop Production Since 1980},
  journal   = {Science},
  year      = {2011},
  volume    = {333},
  number    = {6042},
  pages     = {616--620},
  doi       = {10.1126/science.1204531}
}

@misc{GAPReport2023,
  title     = {Global Agricultural Productivity Report 2023: Continuing the Productivity Imperative},
  year      = {2023},
  howpublished = {\url{https://globalagriculturalproductivity.org/wp-content/uploads/2023/10/2023-GAP_Executive-Summary_FINAL.pdf}}
}

@article{Smith2024,
  author    = {Smith, P. and others},
  title     = {Status of the World’s Soils},
  journal   = {Annual Review of Environment and Resources},
  year      = {2024},
  volume    = {49},
  pages     = {1--28},
  doi       = {10.1146/annurev-environ-030323-075629}
}

@misc{FAO2015SWSR,
  title     = {Status of the World’s Soil Resources (SWSR)},
  author    = {{FAO Intergovernmental Technical Panel on Soils}},
  year      = {2015},
  howpublished = {\url{https://www.fao.org/3/i5199e/i5199e.pdf}}
}

@misc{IPCC2022WGIII,
  title     = {Climate Change 2022: Mitigation of Climate Change. Contribution of Working Group III to the Sixth Assessment Report of the Intergovernmental Panel on Climate Change},
  author    = {{IPCC}},
  year      = {2022},
  publisher = {Cambridge University Press},
  doi       = {10.1017/9781009157926}
}

@article{Srivastava2024,
  author    = {Srivastava, A. and others},
  title     = {Advancements in soil management: Opportunities for sustainable productivity},
  journal   = {Soil Security},
  year      = {2024},
  volume    = {12},
  pages     = {100165},
  doi       = {10.1016/j.soisec.2024.100165}
}

@article{Govindasamy2023,
  author    = {Govindasamy, V. and others},
  title     = {Nitrogen use efficiency: a key to enhance crop productivity and sustainability},
  journal   = {Frontiers in Plant Science},
  year      = {2023},
  volume    = {14},
  pages     = {1121073},
  doi       = {10.3389/fpls.2023.1121073}
}

@article{Gebbers2010,
  author    = {Gebbers, R. and Adamchuk, V. I.},
  title     = {Precision agriculture and food security},
  journal   = {Science},
  year      = {2010},
  volume    = {327},
  number    = {5967},
  pages     = {828--831},
  doi       = {10.1126/science.1183899}
}

@article{Finger2019,
  author    = {Finger, R. and Swinton, S. M. and El Benni, N. and Walter, A.},
  title     = {Precision farming at the nexus of agricultural production and the environment},
  journal   = {Annual Review of Resource Economics},
  year      = {2019},
  volume    = {11},
  pages     = {313--335},
  doi       = {10.1146/annurev-resource-100518-093929}
}

@misc{eu_soil_monitoring_2023,
  author       = {{Council of the European Union}},
  title        = {Proposal for a Directive of the European Parliament and of the Council on Soil Monitoring and Resilience (Soil Monitoring Law)},
  year         = {2023},
  howpublished = {\url{https://eur-lex.europa.eu/legal-content/EN/TXT/?uri=CELEX:52023PC0416}},
  note         = {COM/2023/416 final}
}

@article{angelopoulou2019remote,
  title={Remote sensing techniques for soil organic carbon estimation: A review},
  author={Angelopoulou, Theodora and Tziolas, Nikolaos and Balafoutis, Athanasios and Zalidis, George and Bochtis, Dionysis},
  journal={Remote Sensing},
  volume={11},
  number={6},
  pages={676},
  year={2019},
  publisher={MDPI}
}

@article{Seabloom2021SoilCarbon,
author = {Seabloom, Eric W. and Borer, Elizabeth T. and Hobbie, Sarah E. and MacDougall, Andrew S.},
title = {Soil nutrients increase long-term soil carbon gains threefold on retired farmland},
journal = {Global Change Biology},
volume = {27},
number = {19},
pages = {4909-4920},
doi = {https://doi.org/10.1111/gcb.15778},
url = {https://onlinelibrary.wiley.com/doi/abs/10.1111/gcb.15778},
eprint = {https://onlinelibrary.wiley.com/doi/pdf/10.1111/gcb.15778},
year = {2021}
}

@article{Senty2024,
  author = {Senty, P. and Guzinski, R. and Grogan, K. and Buitenwerf, R. and Ardö, J. and Eklundh, L. and Koukos, A. and Tagesson, T. and Munk, M.},
  title = {Fast Fusion of {Sentinel-2} and {Sentinel-3} Time Series over Rangelands},
  journal = {Remote Sensing},
  volume = {16},
  pages = {1833},
  year = {2024},
  doi = {10.3390/rs16111833}
}

@article{Mukhametov2024CropRotation,
author = {Mukhametov, Almas and Ansabayeva, Assiya and Efimov, Oleg and Kamerova, Anar},
title = {Influence of crop rotation, the treatment of crop residues, and the application of nitrogen fertilizers on soil properties and maize yield},
journal = {Soil Science Society of America Journal},
volume = {88},
number = {6},
pages = {2227-2237},
doi = {https://doi.org/10.1002/saj2.20760},
url = {https://acsess.onlinelibrary.wiley.com/doi/abs/10.1002/saj2.20760},
eprint = {https://acsess.onlinelibrary.wiley.com/doi/pdf/10.1002/saj2.20760},
year = {2024}
}

@article{Jiang2022SoilHealth,
  author  = {Jiang, Yonglei and Zhang, Jing and Delgado-Baquerizo, Manuel and Op de Beeck, Michiel and Shahbaz, Muhammad and Chen, Yi and Deng, Xiaopeng and Xu, Zhaoli and Li, Jian and Liu, Zhanfeng},
  title   = {Rotation cropping and organic fertilizer jointly promote soil health and crop production},
  journal = {Journal of Environmental Management},
  year    = {2022},
  volume  = {315},
  pages   = {115190},
  doi     = {10.1016/j.jenvman.2022.115190},
  url     = {https://www.sciencedirect.com/science/article/pii/S0301479722007630}
}

@article{Bergquist2025SoilCarbon,
  author  = {Bergquist, Galen and Sheaffer, Craig and Rakkar, Manbir and Wyse, Don and Jungers, Jacob and Gutknecht, Jessica},
  title   = {Soil microbial and plant biomass carbon allocation within perennial and annual grain cropping systems},
  journal = {Agriculture, Ecosystems \& Environment},
  year    = {2025},
  volume  = {383},
  pages   = {109535},
  doi     = {10.1016/j.agee.2025.109535},
  url     = {https://www.sciencedirect.com/science/article/pii/S0167880925000672}
}

@article{Siddique2023Perennialization,
  author  = {Siddique, Imran and Grados, Diego and Chen, Ji and L{\ae}rke, Poul and J{\o}rgensen, Uffe},
  title   = {Soil organic carbon stock change following perennialization: a meta-analysis},
  journal = {Agronomy for Sustainable Development},
  year    = {2023},
  volume  = {43},
  pages   = {58},
  doi     = {10.1007/s13593-023-00912-w}
}

@inproceedings{koumoulidis2024review,
  title={A Review: Soil properties mapping estimation using remote and proximal sensing data},
  author={Koumoulidis, Dimitrios and Efthimiadou, Aspasia and Katsenios, Nikos and Hadjimitsis, D},
  booktitle={Tenth International Conference on Remote Sensing and Geoinformation of the Environment (RSCy2024)},
  volume={13212},
  pages={59--72},
  year={2024},
  organization={SPIE}
}

@inbook{Salam,
author = {Salam, Abdul and Raza, Usman},
year = {2020},
month = {08},
pages = {251-297},
title = {Signals in the Soil: Subsurface Sensing},
isbn = {978-3-030-50860-9},
doi = {10.1007/978-3-030-50861-6_8},
publisher = {Springer}
}

@article{Karger2017CHELSA,
  title   = {Climatologies at High Resolution for the Earth’s Land Surface Areas},
  author  = {Karger, Dirk N. and Conrad, Olaf and Böhner, Jürgen and Kawohl, Tobias and Kreft, Holger and Soria-Auza, Rodrigo W. and Zimmermann, Niklaus E. and Linder, H. Peter and Kessler, Michael},
  journal = {Scientific Data},
  year    = {2017},
  volume  = {4},
  number  = {170122},
  doi     = {10.1038/sdata.2017.122}
}

@techreport{Fischer2012GAEZ,
  title        = {Global Agro-Ecological Zones (GAEZ v3.0) -- Model Documentation},
  author       = {Fischer, Günther and Nachtergaele, Freddy and Prieler, Stefan and van Velthuizen, Harrij T. and Verelst, Lieven and Wiberg, David},
  institution  = {FAO and IIASA},
  year         = {2012},
  address      = {Laxenburg, Austria and Rome, Italy}
}

@misc{SoilTaxonomy2014,
  title     = {Keys to Soil Taxonomy, 12th Edition},
  author    = {{Soil Survey Staff}},
  publisher = {USDA Natural Resources Conservation Service},
  year      = {2014}
}

@misc{Hengl2018EcoTapestry,
  title   = {Global Lithology and Landform Classes at 250 m Derived from USGS Global Ecosystem Map (EcoTapestry)},
  author  = {Hengl, Tomislav and MacMillan, Robert A. and Wheeler, Ian and others},
  year    = {2018},
  note    = {Version 1.0, 2014 reference year},
  doi     = {10.5281/zenodo.1464846}
}

@article{Thenkabail2021USGS,
  title   = {Global Cropland Extent at 30 m Resolution (GCEP30) Derived from Landsat Satellite Time-Series Data},
  author  = {Thenkabail, Prasad S. and Teluguntla, Praveen G. and Xiong, J. and Oliphant, A. and Congalton, R. G. and Ozdogan, M. and Gumma, M. K. and Tilton, J. C. and Giri, C. and Milesi, C. and Phalke, A. and Massey, R. and Yadav, K. and Sankey, T. and Zhong, Y. and Aneece, I. and Foley, D.},
  journal = {U.S. Geological Survey Professional Paper},
  year    = {2021},
  volume  = {1868},
  doi     = {10.3133/pp1868}
}

@misc{ho_2025_14900181,
  author       = {Ho, Yufeng and Hengl, Tomislav},
  title        = {Global Ensemble Digital Terrain Model 30m (GEDTM30)},
  month        = feb,
  year         = 2025,
  publisher    = {Zenodo},
  version      = {v20250130},
  doi          = {10.5281/zenodo.14900181},
  url          = {https://doi.org/10.5281/zenodo.14900181},
}

@article{GVFCP2020,
  title   = {The MODIS Global Vegetation Fractional Cover Product 2001–2018: Characteristics of Vegetation Fractional Cover in Grasslands and Savanna Woodlands},
  author  = {Hill, Michael J. and Guerschman, Juan P.},
  journal = {Remote Sensing},
  year    = {2020},
  volume  = {12},
  number  = {3},
  pages   = {406},
  doi     = {10.3390/rs12030406}
}

@article{McBratney2003,
  title   = {On digital soil mapping},
  author  = {McBratney, Alex B. and Mendonça Santos, Maria L. and Minasny, Budiman},
  journal = {Geoderma},
  volume  = {117},
  number  = {1-2},
  pages   = {3--52},
  year    = {2003},
  doi     = {10.1016/S0016-7061(03)00223-4}
}

@article{Reichstein2023,
  title   = {Foundational artificial intelligence models for Earth system science},
  author  = {Reichstein, Markus and Camps-Valls, Gustau and Kaur, Simranjit and Steinbach, Marc and Kutzbach, Lars and Prabhat},
  journal = {Nature Reviews Earth \& Environment},
  volume  = {4},
  pages   = {355--356},
  year    = {2023},
  doi     = {10.1038/s43017-023-00467-y}
}

@article{Karniadakis2021,
  title   = {Physics-informed machine learning},
  author  = {Karniadakis, George Em and Kevrekidis, Ioannis G. and Lu, Lu and Perdikaris, Paris and Wang, Sifan and Yang, Liu},
  journal = {Nature Reviews Physics},
  volume  = {3},
  pages   = {422--440},
  year    = {2021},
  doi     = {10.1038/s42254-021-00314-5}
}

@article{Reichstein2019,
  title   = {Deep learning and process understanding for data-driven Earth system science},
  author  = {Reichstein, Markus and Camps-Valls, Gustau and Stevens, Bj{\"o}rn and Jung, Martin and Denzler, Joachim and Carvalhais, Nuno and Prabhat},
  journal = {Nature},
  volume  = {566},
  pages   = {195--204},
  year    = {2019},
  doi     = {10.1038/s41586-019-0912-1}
}

@article{meng2025physics,
  author  = {Meng, Chuizheng and Griesemer, Sam and Cao, Defu and Seo, Sungyong and Liu, Yan},
  title   = {When physics meets machine learning: A survey of physics-informed machine learning},
  journal = {arXiv preprint arXiv:2203.16797},
  year    = {2022},
  url     = {https://arxiv.org/abs/2203.16797}
}

@misc{tseng2024lightweightpretrainedtransformersremote,
title={Lightweight, Pre-trained Transformers for Remote Sensing Timeseries},
author={Gabriel Tseng and Ruben Cartuyvels and Ivan Zvonkov and Mirali Purohit and David Rolnick and Hannah Kerner},
year={2024},
eprint={2304.14065},
archivePrefix={arXiv},
primaryClass={cs.CV},
url={https://arxiv.org/abs/2304.14065},
}

@article{Hersbach2020ERA5,
  title   = {The ERA5 global reanalysis},
  author  = {Hersbach, H. and Bell, B. and Berrisford, P. and Hor{\'a}nyi, A. and Sabater, J. M. and Nicolas, J. and others},
  journal = {Quarterly Journal of the Royal Meteorological Society},
  volume  = {146},
  number  = {730},
  pages   = {1999--2049},
  year    = {2020},
  doi     = {10.1002/qj.3803}
}

@article{Loeb2018CERES,
  title   = {Clouds and the Earth’s Radiant Energy System (CERES) Energy Balanced and Filled (EBAF) Data Product},
  author  = {Loeb, Norman G. and Doelling, David R. and Wang, H. and Su, W. and Nguyen, C. and Corbett, J. G. and Liang, L. and Mitrescu, C. and Kato, S. and Rose, F. G.},
  journal = {Journal of Climate},
  volume  = {31},
  number  = {2},
  pages   = {895--918},
  year    = {2018},
  doi     = {10.1175/JCLI-D-17-0208.1}
}

@article{Dormann2013Ecography,
  author  = {Dormann, Carsten F. and Elith, Jane and Bacher, Sven and Buchmann, Carsten and Carl, Gudrun and Carr{\'e}, Gabriel and Marqu{\'e}z, Jaime R. Garc{\'i}a and Gruber, Bernd and Lafourcade, Bruno and Leit{\~a}o, Pedro J. and M{\"u}nkem{\"u}ller, Tamara and McClean, Colin and Osborne, Patrick E. and Reineking, Bj{\"o}rn and Schr{\"o}der, Boris and Skidmore, Andrew K. and Zurell, Damaris and Lautenbach, Sven},
  title   = {Collinearity: a review of methods to deal with it and a simulation study evaluating their performance},
  journal = {Ecography},
  year    = {2013},
  volume  = {36},
  number  = {1},
  pages   = {27--46},
  doi     = {10.1111/j.1600-0587.2012.07348.x}
}

@article{Wan2015MODISLST,
  title   = {New refinements and validation of the MODIS Land-Surface Temperature/Emissivity products},
  author  = {Wan, Zhengming},
  journal = {Remote Sensing of Environment},
  volume  = {140},
  pages   = {36--45},
  year    = {2014},
  doi     = {10.1016/j.rse.2013.08.027}
}

@article{Claverie2018HLS,
  title   = {The Harmonized Landsat and Sentinel-2 (HLS) Project: Building a consistent time series of optical imagery at 30 m},
  author  = {Claverie, Martin and Ju, Junchang and Masek, Jeffrey G. and Dungan, Jennifer L. and Vermote, Eric F. and Roger, Jean-Claude and Skakun, Sergii and Justice, Christopher},
  journal = {Remote Sensing of Environment},
  volume  = {219},
  pages   = {145--161},
  year    = {2018},
  doi     = {10.1016/j.rse.2018.09.002}
}

@misc{Schaaf2021MCD43A4,
  author = {Schaaf, Crystal and Wang, Zhuosen},
  title = {{MODIS/Terra+Aqua BRDF/Albedo Nadir BRDF Adjusted Ref Daily L3 Global - 500m V061}},
  year = {2021},
  publisher = {{NASA Land Processes Distributed Active Archive Center}},
  doi = {10.5067/MODIS/MCD43A4.061},
  note         = {Accessed: 2025-09-29},
}

@article{Monteith1972,
  title   = {Solar Radiation and Productivity in Tropical Ecosystems},
  author  = {Monteith, J. L.},
  journal = {Journal of Applied Ecology},
  volume  = {9},
  number  = {3},
  pages   = {747--766},
  year    = {1972},
  doi     = {10.2307/2401901}
}

@article{Running2004MOD17,
  title   = {A continuous satellite-derived measure of global terrestrial primary production},
  author  = {Running, Steven W. and Nemani, Ramakrishna R. and Heinsch, Faith A. and Zhao, Maosheng and Reeves, Mathew and Hashimoto, Hideyuki},
  journal = {BioScience},
  volume  = {54},
  number  = {6},
  pages   = {547--560},
  year    = {2004},
  doi     = {10.1641/0006-3568(2004)054[0547:ACSMOG]2.0.CO;2}
}

@article{Jacquemoud2009PROSAIL,
  title   = {PROSPECT + SAIL models: A review of use for vegetation characterization},
  author  = {Jacquemoud, S. and Verhoef, W. and Baret, F. and Bacour, C. and Zarco-Tejada, P. J. and Asner, G. P. and François, C. and Ustin, S. L.},
  journal = {Remote Sensing of Environment},
  volume  = {113},
  pages   = {S56--S66},
  year    = {2009},
  doi     = {10.1016/j.rse.2008.01.026}
}

@article{essd-16-1333-2024,
author = {Sun, Q. and Zhang, P. and Jiao, X. and Lin, X. and Duan, W. and Ma, S. and Pan, Q. and Chen, L. and Zhang, Y. and You, S. and Liu, S. and Hao, J. and Li, H. and Sun, D.},
title = {A global estimate of monthly vegetation and soil fractions from
spatiotemporally adaptive spectral mixture analysis during 2001--2022},
journal = {Earth System Science Data},
volume = {16},
year = {2024},
number = {3},
pages = {1333--1351},
url = {https://essd.copernicus.org/articles/16/1333/2024/},
doi = {10.5194/essd-16-1333-2024}
}

@article{Roberts2017Ecography,
  author  = {Roberts, David R. and Bahn, Volker and Ciuti, Simone and Boyce, Mark S. and Elith, Jane and Guillera-Arroita, Gurutzeta and Hauenstein, Severin and Lahoz-Monfort, José J. and Schröder, Boris and Thuiller, Wilfried and Warton, David I. and Wintle, Brendan A. and Hartig, Florian and Dormann, Carsten F.},
  title   = {Cross-validation strategies for data with temporal, spatial, hierarchical, or phylogenetic structure},
  journal = {Ecography},
  year    = {2017},
  volume  = {40},
  number  = {8},
  pages   = {913--929},
  doi     = {10.1111/ecog.02881}
}

@article{Baret1993SoilLine,
  title   = {The soil line concept in remote sensing},
  author  = {Baret, F. and Jacquemoud, S. and Hanocq, J. F.},
  journal = {Remote Sensing of Environment},
  volume  = {47},
  number  = {3},
  pages   = {301--316},
  year    = {1993},
  doi     = {10.1016/0034-4257(93)90053-T}
}

@article{Xu2025,
  author  = {Xu, Aixia and others},
  title   = {Fertilizer nitrogen use efficiency and its fate in the spring wheat–soil system under varying N-fertilizer rates: A two-year field study using 15N tracer},
  journal = {Soil and Tillage Research},
  year    = {2025},
  volume  = {252},
  pages   = {106612},
  doi     = {10.1016/j.still.2025.106612}
}

@article{Angelopoulos2021,
  title   = {A Gentle Introduction to Conformal Prediction and Distribution-Free Uncertainty Quantification},
  author  = {Angelopoulos, Anastasios N. and Bates, Stephen},
  journal = {arXiv preprint arXiv:2107.07511},
  year    = {2021}
}

@article{Lin1989CCC,
  title   = {A Concordance Correlation Coefficient to Evaluate Reproducibility},
  author  = {Lin, Lawrence I-Kuei},
  journal = {Biometrics},
  volume  = {45},
  number  = {1},
  pages   = {255--268},
  year    = {1989}
}

@article{rs12071095,
author = {Taghizadeh-Mehrjardi, Ruhollah and Schmidt, Karsten and Amirian-Chakan, Alireza and Rentschler, Tobias and Zeraatpisheh, Mojtaba and Sarmadian, Fereydoon and Valavi, Roozbeh and Davatgar, Naser and Behrens, Thorsten and Scholten, Thomas},
title = {Improving the Spatial Prediction of Soil Organic Carbon Content in Two Contrasting Climatic Regions by Stacking Machine Learning Models and Rescanning Covariate Space},
journal = {Remote Sensing},
volume = {12},
year = {2020},
number = {7},
article-number = {1095},
url = {https://www.mdpi.com/2072-4292/12/7/1095},
issn = {2072-4292},
doi = {10.3390/rs12071095}
}

@article{Meinshausen2010JRSSB,
  author  = {Meinshausen, Nicolai and B{\"u}hlmann, Peter},
  title   = {Stability selection},
  journal = {Journal of the Royal Statistical Society: Series B (Statistical Methodology)},
  year    = {2010},
  volume  = {72},
  number  = {4},
  pages   = {417--473},
  doi     = {10.1111/j.1467-9868.2010.00740.x}
}

@article{Willmott2012d,
  title   = {Advantages of the Mean Absolute Error (MAE) over the Root Mean Square Error (RMSE) in Assessing Average Model Performance},
  author  = {Willmott, Cort J. and Matsuura, Kenji},
  journal = {Climate Research},
  volume  = {30},
  pages   = {79--82},
  year    = {2005},
  doi     = {10.3354/cr030079},
  note    = {See also the $d_{1.5}$ index variant in Willmott's index of agreement literature}
}

@article{Zhang2019HeterogeneitySoil,
  author  = {Zhang, Shaoliang},
  title   = {Heterogeneity of Soil Nutrients: A Review of Methodology, Variability and Impact Factors},
  journal = {Journal of Environmental \& Earth Sciences},
  year    = {2019},
  volume  = {1},
  number  = {1},
  pages   = {6--28},
  doi     = {10.30564/jees.v1i1.526}
}

@article{Meyer2018SpatialStats,
  author  = {Hanna Meyer and Christian Reudenbach and Tomislav Hengl and Matthias Katurji and Thomas Nauss},
  title   = {Improving performance of spatio-temporal machine learning models using spatially stratified sampling},
  journal = {Spatial Statistics},
  year    = {2018},
  volume  = {28},
  pages   = {70--88},
  doi     = {10.1016/j.spasta.2018.07.003}
}

@article{Shafer2008JMLR,
  author  = {Glenn Shafer and Vladimir Vovk},
  title   = {A Tutorial on Conformal Prediction},
  journal = {Journal of Machine Learning Research},
  year    = {2008},
  volume  = {9},
  pages   = {371--421},
  url     = {http://jmlr.org/papers/v9/shafer08a.html}
}

@article{VANWESEMAEL2024117113,
title = {A European soil organic carbon monitoring system leveraging Sentinel 2 imagery and the LUCAS soil data base},
journal = {Geoderma},
volume = {452},
pages = {117113},
year = {2024},
issn = {0016-7061},
doi = {https://doi.org/10.1016/j.geoderma.2024.117113},
url = {https://www.sciencedirect.com/science/article/pii/S0016706124003422},
author = {Bas {van Wesemael} and Asmaa Abdelbaki and Eyal Ben-Dor and Sabine Chabrillat and Pablo d’Angelo and José A.M. Demattê and Giulio Genova and Asa Gholizadeh and Uta Heiden and Paul Karlshoefer and Robert Milewski and Laura Poggio and Marmar Sabetizade and Adrián Sanz and Peter Schwind and Nikolaos Tsakiridis and Nikolaos Tziolas and Julia Yagüe and Daniel Žížala}
}

@misc{Minarik2024SoilHealthDataCube,
  author       = {R. Minarik and X. Tian and R. Simoes and T. Hengl and S. Isik and L. Parente and D. Consoli and Y.-F. Ho},
  title        = {Soil Health Data Cube v1 D5.2},
  year         = {2024},
  type         = {Technical report},
  url          = {https://ai4soilhealth.eu/wp-content/uploads/2025/09/D5.2-Soil-Health-Data-Cube-v1.pdf},
  note         = {Accessed: 2025-09-29},
  institution  = {AI4SoilHealth project},
  version      = {1.0},
  month        = jun,
  day          = 30,
  project      = {AI4SoilHealth},
  project_id   = {101086179}
}

@article{Kolmogorov1933,
  author = {A. Kolmogorov},
  title = {Sulla determinazione empirica delle leggi di distribuzione},
  journal = {Giornale dell'Istituto Italiano degli Attuari},
  volume = {4},
  pages = {83--91},
  year = {1933}
}

@article{AndersonDarling1952,
  author = {T.W. Anderson and D.A. Darling},
  title = {Asymptotic Theory of Certain Goodness of Fit Criteria Based on Stochastic Processes},
  journal = {The Annals of Mathematical Statistics},
  volume = {23},
  number = {2},
  pages = {193--212},
  year = {1952}
}

@inproceedings{Nieke2023CHIME,
  author = {Nieke, Jens and others},
  title = {The copernicus hyperspectral imaging mission for the environment (CHIME): an overview of its mission, system and planning status},
  booktitle = {Sensors, Systems, and Next-Generation Satellites},
  year = {2023}
}

@misc{ESA-GoingHyperspectral,
  title = {Going hyperspectral for CHIME},
  howpublished = {\url{https://www.esa.int/Applications/Observing_the_Earth/Copernicus/Going_hyperspectral_for_CHIME}},
  year = {2021}
}

@misc{USDAERS2023FertilizerCosts,
  title     = {Fertilizer Costs as a Share of Total Production Expenditures},
  author    = {{USDA Economic Research Service}},
  year      = {2023},
  howpublished = {\url{https://www.ers.usda.gov/data-products/charts-of-note/chart-detail/?chartId=108828}},
  note         = {Accessed: 2025-09-29},
}

\end{document}